\pdfoutput=1
\documentclass[10pt,twocolumn,letterpaper]{article}

\usepackage[letterpaper,top=1in,left=0.8125in,textwidth=6.875in,textheight=8.875in,columnsep=0.3125in,headheight=0pt,headsep=0pt,footskip=30pt]{geometry}
\usepackage{times}
\usepackage[dvipsnames]{xcolor}
\usepackage{graphicx}
\usepackage{amsmath}
\usepackage{amssymb}
\usepackage{booktabs}
\usepackage[numbers,sort&compress]{natbib}
\usepackage[format=plain,labelformat=simple,labelsep=period,font=small,compatibility=false]{caption}
\usepackage[font=footnotesize,skip=3pt,subrefformat=parens]{subcaption}
\usepackage[hyphens]{url}
\usepackage[shortlabels,inline]{enumitem}
\setlist[itemize]{noitemsep,leftmargin=*,topsep=0em}
\setlist[enumerate]{noitemsep,leftmargin=*,topsep=0em}

\makeatletter
\renewcommand\section{\@startsection{section}{1}{\z@}{-10pt plus -2pt minus -2pt}{7pt}{\large\bfseries}}
\renewcommand\subsection{\@startsection{subsection}{2}{\z@}{-8pt plus -2pt minus -2pt}{5pt}{\normalsize\bfseries}}

\newcommand{\papertitle}{NEST: Uncovering and Nesting Unknown Objects\\in Open-World Object Detection}
\title{\papertitle}
\author{Yuchen Zhang \qquad Yuan Gao \qquad Sebastian Schmidt \qquad Johannes Betz\\
Technical University of Munich, Germany\\
{\small\ttfamily yuchen2.zhang@tum.de}}
\date{}
\renewcommand{\@maketitle}{%
  \newpage\null\vskip 0pt
  \begin{center}
    {\Large\bfseries \@title\par}
    \vskip 20pt
    {\large\lineskip .5em
      \begin{tabular}[t]{c}\@author\end{tabular}\par}
    \vskip 20.103pt
  \end{center}}
\renewenvironment{abstract}{\section*{Abstract}\leftskip=0.25pc\rightskip=0.25pc}{\par\vspace{13.66pt}}
\newcommand{\makesupptitle}{%
  \newpage
  \twocolumn[\centering\Large\textbf{\papertitle}\\\vspace{0.5em}Supplementary Material\\\vspace{1.0em}]}
\newcommand{\etal}{\emph{et al}.\null}
\makeatother

\usepackage{colortbl}
\usepackage{multirow}
\colorlet{oursgray}{gray!14}
\definecolor{oursblue}{HTML}{E7EFF7}

\definecolor{linkblue}{rgb}{0.21,0.49,0.74}
\usepackage[breaklinks,colorlinks,allcolors=linkblue]{hyperref}
\usepackage[acronym,nomain]{glossaries}
\usepackage[capitalize]{cleveref}
\crefname{section}{Sec.}{Secs.}
\Crefname{section}{Section}{Sections}
\crefname{table}{Tab.}{Tabs.}
\Crefname{table}{Table}{Tables}

\glsdisablehyper
\setacronymstyle{long-short}
\newacronym{owod}{OWOD}{open-world object detection}
\newacronym{ovd}{OVD}{open-vocabulary detection}
\newacronym{iou}{IoU}{intersection over union}
\newacronym{nms}{NMS}{non-maximum suppression}
\newacronym{ap}{AP}{average precision}
\newacronym{map}{mAP}{mean average precision}
\newacronym{urec}{U-Rec}{unknown recall}
\newacronym{auc}{AUC}{area under the curve}
\newacronym{e2e}{E2E}{end-to-end recall}
\newacronym{aose}{A-OSE}{absolute open-set error}
\newacronym{svm}{SVM}{support vector machine}

\begin{document}
\maketitle
\begin{abstract}
Open-world object detectors aim to localize objects of classes not covered by their training annotations, but most label them only as \emph{unknown}.
This label tells a system that an object is present but not what it is, while its broader category would allow a more informed decision.
We therefore address both uncovering unknown objects and assigning them to the broader categories of a given class taxonomy.
For uncovering, we find that most unknown objects are already localized and labeled \emph{unknown} in our base detector, but are discarded during final top-$k$ selection because background receives higher scores.
To mitigate this, we propose a class-agnostic ranker that estimates how well each predicted box fits an object and uses these scores to select the final predictions.
Across three benchmarks, our detector improves unknown recall by 3.8\% to 195.1\% relative to the strongest baseline in each task, while maintaining competitive known-class detection.
For categorizing, we draw on Euler diagrams, in which the region of a category contains the regions of its members.
Our categorizer builds each category's region from the known classes and learns how much room to leave for members it has never seen.
On the same detections, it categorizes unknown objects more accurately than existing hierarchical methods.
Code is available at \url{https://github.com/TUM-AVS/NEST}.
\end{abstract}

\vspace{-15pt}
\section{Introduction}
\label{sec:intro}

Detectors deployed in the real world may encounter objects outside their annotated training classes.
\Gls{owod}~\cite{joseph2021towards} aims to localize such objects and label them as \emph{unknown}, while still recognizing the known classes.
This label, however, treats all unknown objects alike. A broader category such as \emph{animal} could help an autonomous system anticipate that an unfamiliar zebra may move on its own, unlike a fallen tree.
We therefore argue that an open-world detector should answer three questions about each detection, namely (1)~\textbf{whether it is an object}, (2)~\textbf{whether it belongs to a known class} and (3)~\textbf{if not, to which broader category it belongs}.

Only question~(2) can be learned directly from known-class annotations.
Questions~(1) and~(3) concern objects of unknown classes, which carry no annotations, so a detector must answer both by generalizing from the known classes.
For question~(1), existing \gls{owod} methods generalize, for example, by pseudo-labeling unannotated regions~\cite{joseph2021towards,gupta2022owdetr} or by learning a class-agnostic objectness score from known objects~\cite{zohar2023prob}.
Question~(3), which we term \emph{unknown object categorization}, has been studied mainly in image classification, for example by HND~\cite{lee2018hierarchical} and ProHOC~\cite{wallin2025prohoc}.
Most \gls{owod} methods stop at the first two questions, and only a few detectors answer all three~\cite{zhou2023unknown,azeem2026hsgdet}.

One might expect that answering question~(1) requires better localization of unknown objects.
We find instead that \textbf{final top-$k$ selection is the dominant bottleneck} in our base detector.
Most unknown objects are already well localized, but their detections are discarded because background receives higher classification scores.
The \textbf{first part of NEST} is therefore a detector built around a class-agnostic ranker trained only on known-class annotations to predict box localization quality~\cite{jiang2018iounet}.
Selecting predictions by this quality rather than classification confidence allows well-localized unknown detections to reach the final output.
The \textbf{second part of NEST}, a categorizer for question~(3), draws on Euler diagrams, in which the region of a broader category, which we call a group, contains the regions of its members (\cref{fig:overview}).
It builds each group's region in feature space from its known classes, such as \emph{dog} and \emph{giraffe} for \emph{animal}, and learns to leave room for unseen members such as the zebra.

We evaluate NEST on three \gls{owod} benchmarks.
In M-OWODB~\cite{joseph2021towards} and S-OWODB~\cite{gupta2022owdetr}, classes within a group are mostly introduced together, leaving nearly all unknown objects with \emph{object} as their target under our protocol.
We therefore construct R-OWODB by introducing COCO classes~\cite{lin2014coco} in random order.
Our detector improves unknown recall by 3.8\% to 195.1\% relative to the strongest baseline in each task, while maintaining competitive known-class detection.
On R-OWODB, our categorizer achieves higher accuracy than four hierarchical baselines.

In summary, we make three contributions.

\begin{itemize}
  \item We identify final top-$k$ selection as the dominant bottleneck in our base detector and introduce a class-agnostic ranker that recovers unknown detections using predicted box localization quality.

  \item We propose a categorizer that builds group distributions from their known classes and learns how much to widen them for unseen members. It is detector-agnostic, and other open-world detectors need no retraining to adopt it.

 \item We evaluate NEST on \emph{three detection benchmarks} and adapt \emph{four hierarchical classification methods} to compare categorizers on the same detections. The results show higher unknown recall and categorization accuracy with competitive known-class detection.
\end{itemize}

\section{Related Work}
\label{sec:related}

\subsection{Open-World Object Detection}
\label{sec:related_owod}

\Gls{owod} asks a detector to localize objects of classes it was never taught, label them as \emph{unknown} and learn these classes later once they are annotated~\cite{joseph2021towards}.
Most \gls{owod} methods start from an ImageNet-pretrained backbone and learn unknown detection from task images annotated only for known classes.
Three complementary directions are to mine unknown pseudo-labels, model objectness probabilistically and adapt proposal generation.
\textbf{The first direction} trains on pseudo-labels for unannotated regions likely to contain objects.
Candidates are selected using proposal scores~\cite{joseph2021towards}, backbone activation~\cite{gupta2022owdetr}, activation cues combined with selective search~\cite{ma2023cat}, hyperbolic similarity to known-class features~\cite{doan2024hypow} or feature dissimilarity to known objects and background~\cite{majee2025crowd}.
\textbf{The second direction} models objectness probabilistically.
PROB~\cite{zohar2023prob} uses a Gaussian in query feature space without mining unknown training examples.
Decoupled PROB~\cite{inoue2025decoupled} estimates objectness at an early decoder layer to reduce interference with class prediction.
OWOBJ~\cite{zhang2025owobj} combines probabilistic objectness modeling with pseudo-label supervision, using an adaptive Gaussian prior and converting objectness scores into soft unknown targets for unmatched queries.
\textbf{The third direction} changes how the detector generates proposals and represents objects.
RandBox~\cite{wang2023randbox} reduces known-class bias with random proposals, and its extensions OrthogonalDet~\cite{sun2024exploring} and DEUS~\cite{heo2026deus} further reduce it by disentangling features.

Additional pretrained models provide another source of supervision for unknown detection.
MEPU~\cite{fang2025mepu} obtains unknown pseudo-labels from FreeSOLO~\cite{wang2022freesolo}, a segmentation model trained without labels on all COCO training images.
SKDF~\cite{ma2025skdf} transfers detections from the vision-language model GLIP~\cite{li2022glip}, prompted with the LVIS~\cite{gupta2019lvis} class names.
OW-Rep~\cite{lee2026owrep} refines unknown boxes with masks from SAM~\cite{kirillov2023segment} and distills features from a vision foundation model into its embeddings.

Some works go one step further and divide the unknown objects into new classes.
They cluster the objects found in a collection of unlabeled images~\cite{zheng2022osodd,rambhatla2021pursuit}, train an extra head that sorts them into a fixed number of clusters~\cite{fomenko2022rncdl} or add a fixed number of unknown classes to an open-world detector~\cite{wu2022ucowod}.
Their clusters are anonymous, and a cluster acquires a meaning only when an evaluator matches it to ground-truth classes afterwards.
NEST instead builds on the premise that semantic knowledge learned from known objects can generalize to unknown members of their groups.
It uses this knowledge to assign each unknown object a named taxonomy group, such as \emph{animal} or \emph{furniture}.

\subsection{Open-Vocabulary Detection}
\label{sec:related_ovd}

\Gls{ovd} also goes beyond a fixed label space, but its setting is different.
An \gls{ovd} detector receives an image together with a text prompt and localizes the objects that the prompt names~\cite{zareian2021ovrcnn,minderer2022owlvit,liu2024groundingdino,cheng2024yoloworld}.
Vision-language models such as CLIP~\cite{radford2021clip} make this possible by embedding images and text in a shared space, so that region features can be matched against the text embeddings of arbitrary class names.
ViLD~\cite{gu2022vild} distills CLIP image embeddings into the region features of a detector, and GLIP~\cite{li2022glip} pretrains a detector to ground prompt phrases in the image.
Since the prompt names the classes, \gls{ovd} suits targets known in advance, as when a user asks to ``find the bottle''.
NEST is an \gls{owod} method, not an \gls{ovd} method.
It never receives the names of the unknown classes and uses neither prompts nor text embeddings.

\subsection{Novel Classes in a Class Hierarchy}
\label{sec:related_hier}

Assigning unfamiliar objects to broader nodes in a class hierarchy, such as \emph{animal} for a zebra, has been studied mainly in image classification.
Most methods choose the node at test time from the scores of classifiers trained on the known classes.
DARTS~\cite{deng2012hedging} maximizes the expected information gain subject to a minimum overall accuracy, HSC~\cite{linderman2023finegrain} descends the hierarchy while a per-node confidence exceeds a threshold set on known classes, and ProHOC~\cite{wallin2025prohoc} minimizes the expected tree distance, using novelty scores from separate networks trained at each depth.
HND~\cite{lee2018hierarchical} instead learns an explicit novel output under every internal node, trained by relabeling images of known classes with an ancestor.
Few detectors give such coarser answers.
Zhou~\etal~\cite{zhou2023unknown} apply the training of HND to the classification head of a detector on driving datasets.
HSGDet~\cite{azeem2026hsgdet} navigates a WordNet~\cite{miller1995wordnet} hierarchy embedded in the CLIP text space to place each unknown object in remote-sensing images under a node.
We evaluate the four classification methods above as categorization baselines on the frozen query features of open-world detectors.

\section{Method}
\label{sec:method}

\begin{figure*}[t]
  \centering
  \includegraphics{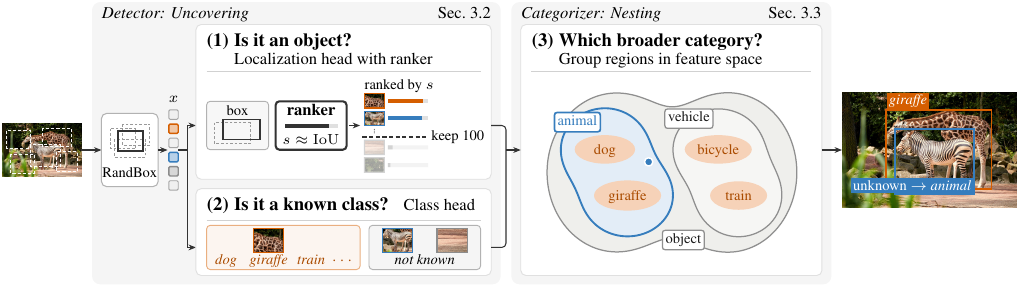}
  \caption{\textbf{Overview of NEST.} Colors mark a known giraffe (orange), an unknown zebra (blue) and background (gray). Our detector uncovers unknown objects by asking two questions of each query feature $x$, and the categorizer nests them by asking a third. (1)~\emph{Is it an object?} The ranker of the localization head predicts how well the refined box fits an object, measured as IoU, and each image keeps the 100 detections it ranks highest. (2)~\emph{Is it a known class?} The class head labels each detection with a known class or as \emph{not known}. (3)~\emph{Which broader category?} The categorizer places the zebra in \emph{animal}, whose region in feature space is built from known classes such as \emph{dog} and \emph{giraffe} and leaves room for unseen members.}
  \label{fig:overview}
\end{figure*}

To make full use of the unknown objects that a detector already localizes, we introduce NEST (\cref{fig:overview}), which consists of a detector built around a class-agnostic \textbf{ranker} (\cref{sec:discovery}) and a \textbf{categorizer}
(\cref{sec:family_geometry}).
The ranker recovers the unknown detections otherwise lost during final selection.
The categorizer, inspired by Euler diagrams, then assigns them to the broader categories of a given class taxonomy.

\subsection{Preliminaries}
\label{sec:problem}

\textbf{OWOD.} A detector in \gls{owod} learns the classes $\mathcal{Y}$ of a dataset over a sequence of tasks without forgetting earlier ones.
Task $t$ introduces new classes $\mathcal{N}_t$, disjoint from the classes $\mathcal{K}_{t-1}$ of earlier tasks, $\mathcal{N}_t\cap\mathcal{K}_{t-1}=\varnothing$.
The known classes are then $\mathcal{K}_t=\mathcal{K}_{t-1}\cup\mathcal{N}_t$, with $\mathcal{K}_0=\varnothing$, and the remaining classes $\mathcal{U}_t=\mathcal{Y}\setminus\mathcal{K}_t$ are unknown.
During training at task $t$, new annotations cover only $\mathcal{N}_t$, and objects of $\mathcal{U}_t$ may appear in the images without annotations \cite{joseph2021towards, gupta2022owdetr}.
At test time, the detector must localize objects, recognize the known classes and label the others as \emph{unknown}.

\textbf{RandBox.} RandBox~\cite{wang2023randbox} builds on Fast R-CNN~\cite{girshick2015fast} with randomly sampled training proposals. It uses the detection head of Sparse R-CNN~\cite{sun2021sparse} to refine proposals over six stages. The final query feature $x$ feeds a localization head and a class head with one output per known class and an \emph{unknown} output. Matched queries receive their annotated class labels. For unknown supervision, RandBox ranks proposals by the sum of their sigmoid scores over the known classes and \emph{unknown}, then assigns \emph{unknown} pseudo-labels to unmatched queries among the top-ranked proposals. The remaining unmatched queries are treated as background.

\subsection{Detector: Uncovering Unknown Objects}
\label{sec:discovery}

We use RandBox without its unknown pseudo-labeling mechanism as our base detector. We repurpose its \emph{unknown} output as \emph{not known} (previously for unknown objects only, now for anything outside the known classes, including background), assigning this label to all unmatched queries.
Tracing the unknown objects of M-OWODB~\cite{joseph2021towards} Task~1 through this detector (\cref{sec:abl-detector}), we find that 61\% of them are already localized and labeled \emph{unknown}, yet lost at the final top-$k$ selection, as background scores higher on \emph{not known}.
We therefore build our detector around a class-agnostic ranker that predicts localization quality to select the final detections.
Unlike the objectness methods of \cref{sec:related_owod}, our detector leaves labeling to the class head and selection to the ranker.

\begin{figure}[htbp]
  \centering
  \includegraphics[width=\columnwidth]{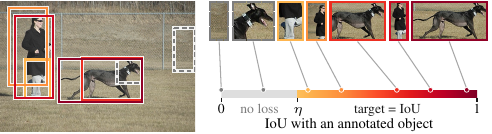}
  \caption{\textbf{How the ranker is trained.} The ranker learns to predict the IoU of a box with the annotated objects whenever this IoU is above $\eta$. The boxes on the woman and the dog are therefore trained toward their IoU, while the box on the background fence and the box covering only the dog's head fall below $\eta$ and are left out of the loss.}
  \label{fig:ranker}
\end{figure}

The ranker is a branch of the localization head that reads the same query feature $x$ and predicts its score $s_i$ for each query $i$ along with the refined box.
For each predicted box $i$, the target $\mathrm{IoU}_i$ is its highest \gls{iou} with any annotated object~\cite{jiang2018iounet,kim2022oln} (\cref{fig:ranker}).
Since unknown objects carry no annotations, boxes on them have a low $\mathrm{IoU}_i$, like boxes on the background.
The loss therefore covers only $\mathcal{P}=\{i:\mathrm{IoU}_i>\eta\}$ for a fixed threshold $\eta\in(0,1)$,
\begin{equation}
  \mathcal{L}_{\mathrm{rank}} = \frac{1}{|\mathcal{P}|} \sum_{i \in \mathcal{P}} \left| s_i - \mathrm{IoU}_i \right|.
  \label{eq:ranker}
\end{equation}
Because \gls{iou} describes how well a box fits an object rather than which object it is, what the ranker learns on known classes carries over to unknown ones.

Most training queries are labeled \emph{not known}.
To account for this imbalance, we follow the logit-adjustment principle of Menon~\etal~\cite{menon2021logit} and subtract $\tau\log(\pi/(1-\pi))$ from the \emph{not known} logit at inference, where $\pi$ is the fraction of training queries assigned to \emph{not known} and $\tau$ controls the adjustment strength.
We select $\tau=0.5$ on held-out training images, as described in Supp.~\cref{sec:supp-impl}, and use this value for all tasks and benchmarks.

Each prediction is then labeled with its highest-scoring known class, or as \emph{unknown} when the \emph{not known} output scores higher.
For class-aware \gls{nms} and evaluation, known predictions are scored by class probability and unknown ones by their ranker score.
The final top-$k$ selection then keeps the $k$ predictions with the highest ranker scores, known and unknown alike.

\subsection{Categorizer: Nesting Unknown Objects}
\label{sec:family_geometry}
\label{sec:hierarchy_training}

\begin{figure}[htbp]
  \centering
  \includegraphics[width=\columnwidth]{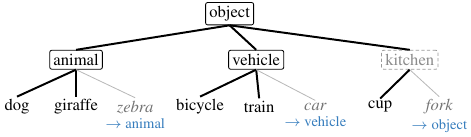}
\caption{\textbf{Categorization targets in our extended OWOD setting.}
Excerpt of $\mathcal{T}$ at Task~1 of R-OWODB (\cref{sec:experiments}).
Known classes are black and unknown classes gray.
Thick edges mark branches containing known classes.
Solid and dashed boxes mark retained and removed groups.
Blue labels give each unknown class's most specific retained group.
With only one known child branch, \emph{kitchen} is removed,
so \emph{fork} targets \emph{object}.}
  \label{fig:taxonomy_example}
\end{figure}

We extend the \gls{owod} setting with
unknown object categorization, assigning each unknown detection
to a broader category in a given taxonomy $\mathcal{T}$. We represent $\mathcal{T}$ as a rooted tree whose root denotes \emph{object}. Its leaves represent object classes, and its internal nodes represent broader categories, which we call \emph{groups}, with classes or more specific groups as children.
At task $t$, the categorizer receives a taxonomy $\mathcal{T}_t$ that retains the known classes $\mathcal{K}_t$, the root and groups with known classes in at least two child branches, since a group whose known classes all lie in one child branch is indistinguishable from that branch~\cite{lee2018hierarchical}.
We denote the retained groups, including the root, by $\mathcal{G}_t$.
For each detection predicted as \emph{unknown}, the categorizer assigns a group $\hat g\in\mathcal{G}_t$.
The target $a_t(u)$ is the most specific retained group containing the object's true class $u\in\mathcal{U}_t$~\cite{wallin2025prohoc}.
\Cref{fig:taxonomy_example} illustrates this with an example.

Inspired by Euler diagrams, our categorizer represents each group as a region in feature space (question~(3) in \cref{fig:overview}).
It constructs these regions from known-class distributions and learns how far they extend beyond known members to accommodate unknown ones.
For example, the \emph{animal} region should accommodate a zebra even when its known members are only dogs and giraffes.

\textbf{Representations.} After the detector is trained, the categorizer learns a projection $z=Px\in\mathbb{R}^d$ of the detector's query features $x$.
The categorizer models each known class $c\in\mathcal{K}_t$ by $p_c(z)=\mathcal{N}(z;\mu_c,\operatorname{diag}(v_c))$ with learned $\mu_c$ and $v_c$.
A group $g$ is modeled by a Gaussian mixture model over its $n_g=|\mathcal{C}(g)|$ immediate children $\mathcal{C}(g)$, with each component widened by a learned variance increment $\Delta_g$ to leave room for unseen members,
\begin{equation}
  p_g(z)=\frac{1}{n_g}\sum_{j\in\mathcal{C}(g)}
  \mathcal{N}\!\left(z;\mu_j,\operatorname{diag}(v_j+\Delta_g)\right).
  \label{eq:family_density}
\end{equation}

When forming the mixture of a parent group, each child group is summarized by a single Gaussian with the mean $\mu_j$ and diagonal variance $v_j$ of its own mixture, which are computed for every group as
\begin{equation}
  \begin{aligned}
    \mu_g&=\frac{1}{n_g}\sum_{j\in\mathcal{C}(g)}\mu_j,\\
    v_g&=\frac{1}{n_g}\sum_{j\in\mathcal{C}(g)}
    \left[v_j+(\mu_j-\mu_g)^{\odot2}\right]+\Delta_g,
  \end{aligned}
  \label{eq:recursive_moments}
\end{equation}
where squaring is elementwise, so the variance combines the variation within children, the spread of their means and the increment $\Delta_g$.
We define the class and group scores as $f_c(z)=\log p_c(z)$ and $f_g(z)=\log p_g(z)$.
At inference, we assign each unknown detection to the retained group with the highest score.

\textbf{Training.}
We train the categorizer on the query features of predictions matched to annotated known objects in the training images.
For a feature $z$ with class label $y\in\mathcal{K}_t$, we use the softmax cross-entropy
\begin{equation}
\ell_{\mathrm{known}}(z,y)
= -\log
\frac{\exp(f_y(z))}
{\sum_{k\in\mathcal{K}_t\cup\mathcal{G}_t}\exp(f_k(z))}.
\label{eq:known_loss}
\end{equation}
This loss encourages the annotated class to score above all other classes and groups.
However, an unknown object has no class model of its own, a case this loss never presents to the categorizer.
Following the leave-one-out strategy of HND~\cite{lee2018hierarchical}, we therefore hide a class or subtree with a cut $C$.
Unlike HND, which removes classifier outputs, the cut in NEST also removes the hidden components from the ancestor mixtures and recomputes their summaries, and a group stays available as long as at least one of its children survives.
Hiding \emph{giraffe}, for example, leaves \emph{animal} with the \emph{dog} component.
The cut loss $\ell_C(z,y)$ is a softmax cross-entropy over the recomputed scores $f_g^C(z)$ of the available groups, with the nearest available ancestor $a_C(y)$ as target. We combine the known-class and cut losses as
\begin{equation}
  \ell_{\mathrm{cat}}(z,y)
  =
  \ell_{\mathrm{known}}(z,y)
  + \sum_C w_C(y)\,\ell_C(z,y),
  \label{eq:categorizer_loss}
\end{equation}
where the sum runs over cuts that hide class $y$, and $w_C(y)$ is the weight of cut $C$ for that class.

\textbf{Cut weighting.}
In the taxonomy of \cref{fig:taxonomy_example}, hiding \emph{giraffe} alone gives the feature of an annotated giraffe the target \emph{animal}, while hiding the \emph{animal} subtree gives it the target \emph{object}.
The first cut imitates an unknown object whose group is retained and the second one whose group is not, a case that becomes rarer in later tasks as more groups are retained.
We therefore weight the cut losses in \cref{eq:categorizer_loss} by how often each case occurs, using $1-\rho_t$ for the first case and $\rho_t$ for the second.
We estimate $\rho_t$ from the known classes as the share of those not directly under the root whose parent group has only two known children, like \emph{animal} for \emph{giraffe}, since such a group would not be retained without the class.
On R-OWODB, $\rho_t$ falls from 0.46 to 0.03 over Tasks~1--3.

\section{Experiments}
\label{sec:experiments}

We evaluate how well NEST uncovers unknown objects (\cref{sec:owod}) and nests them in groups (\cref{sec:categorization}), and show qualitative results in \cref{sec:qualitative}.
The implementation details of our model can be found in Supp.~\cref{sec:supp-impl}.

We build the taxonomy $\mathcal{T}$ from COCO~\cite{lin2014coco}, with the root \emph{object}, the 12 COCO supercategories as groups and the 80 classes as leaves.
The two standard benchmarks, M-OWODB~\cite{joseph2021towards} and S-OWODB~\cite{gupta2022owdetr}, introduce the 80 COCO classes over four tasks.
Since both benchmarks introduce most supercategories within a single task, the target $a_t(u)$ is the root \emph{object} for 90--97\% of unknown test objects in M-OWODB Tasks~1--3 and for 100\% of them in S-OWODB (Supp.~\cref{sec:supp-splits}).
We therefore construct R-OWODB, which splits the 80 COCO classes at random into four tasks of 20, so most supercategories are spread over several tasks.
The share of unknown test objects whose target is the root then falls from 54\% at Task~1 to 17\% at Task~2 and 0\% at Task~3.
All three benchmarks follow the protocol of ORE~\cite{joseph2021towards}, in which each task after the first ends with fine-tuning on a small balanced set of images (Supp.~\cref{sec:supp-splits}).

\subsection{Uncovering Unknown Objects}
\label{sec:owod}

Following prior work~\cite{gupta2022owdetr,zohar2023prob,sun2024exploring}, we report known-class \gls{map} at an \gls{iou} of 0.5 and \gls{urec}, the fraction of unknown test objects matched by a prediction labeled \emph{unknown} at the same \gls{iou}.
We build on RandBox~\cite{wang2023randbox}, so RandBox and its extension OrthogonalDet~\cite{sun2024exploring} are our most direct comparisons.
On M-OWODB and S-OWODB, we take all baseline results from their publications, and on R-OWODB, we train RandBox and OrthogonalDet with their official implementations.

\textbf{Results.}
Our detector achieves the highest \gls{urec} across Tasks~1--3 on all three benchmarks, with relative gains of 3.8\% to 195.1\% over the strongest baseline in each task (\cref{tab:main}).
The gain is smallest on S-OWODB Task~2, where PROB+OWOBJ recalls 28.7\% of the unknown test objects against our 29.8\%, and largest on R-OWODB Task~1, where OrthogonalDet recalls 14.4\% against our 42.5\%.
This range comes mainly from the strongest baseline, whose \gls{urec} varies from 13.2\% to 31.9\% across tasks, while ours stays between 29.8\% and 43.5\%.
Part of our gain also comes from spending more of the shared budget (at most 100 predictions per image) on unknown predictions, but with the same number of unknown predictions as each baseline, our detector still recalls 1.2 to 1.6 times as many unknown objects (Supp.~\cref{sec:supp-quant}).
Known-class detection remains competitive.
On M-OWODB, our detector has the highest \gls{map} in Tasks~1--3 and matches OrthogonalDet in Task~4.
On S-OWODB and R-OWODB, it reaches the highest or joint-highest \gls{map} in three of eight tasks and falls no more than 3.3 points below the best result in the others.
More detailed results, including the absolute open-set error, are in Supp.~\cref{sec:supp-quant}.

\begin{table}[htbp]
\centering
\footnotesize
\setlength{\tabcolsep}{1.94pt}
\renewcommand{\arraystretch}{1.04}
\resizebox{\columnwidth}{!}{%
\begin{tabular}{lccccccc}
\toprule
 & \multicolumn{2}{c}{Task 1} & \multicolumn{2}{c}{Task 2} & \multicolumn{2}{c}{Task 3} & Task 4 \\
\cmidrule(lr){2-3}\cmidrule(lr){4-5}\cmidrule(lr){6-7}\cmidrule(l){8-8}
Detector & mAP$\uparrow$ & U-Rec$\uparrow$ & mAP$\uparrow$ & U-Rec$\uparrow$ & mAP$\uparrow$ & U-Rec$\uparrow$ & mAP$\uparrow$ \\
\midrule
\multicolumn{8}{l}{\textbf{M-OWODB}} \\
OW-DETR~\cite{gupta2022owdetr} & 59.2 & 7.5 & 42.9 & 6.2 & 30.8 & 5.7 & 27.8 \\
UC-OWOD~\cite{wu2022ucowod} & 50.7 & 2.4 & 31.8 & 3.4 & 24.6 & 8.7 & 23.1 \\
PROB~\cite{zohar2023prob} & 59.5 & 19.4 & 44.0 & 17.4 & 36.0 & 19.6 & 31.5 \\
CAT~\cite{ma2023cat} & 60.0 & 23.7 & 44.1 & 19.1 & 34.8 & 24.4 & 29.9 \\
RandBox~\cite{wang2023randbox} & \underline{61.8} & 10.6 & 45.3 & 6.3 & 39.4 & 7.8 & \underline{35.4} \\
Hyp-OW~\cite{doan2024hypow} & 59.4 & 23.5 & 44.4 & 20.6 & 36.8 & 26.3 & 33.6 \\
OrthogonalDet~\cite{sun2024exploring} & 61.3 & \underline{24.6} & \underline{47.0} & \underline{26.3} & \underline{41.3} & \underline{29.1} & \textbf{37.9} \\
Decoupled PROB~\cite{inoue2025decoupled} & 59.8 & 20.3 & 46.6 & 18.4 & 38.5 & 20.3 & 33.8 \\
PROB+OWOBJ~\cite{zhang2025owobj} & 61.4 & 23.6 & 45.7 & 23.8 & 38.8 & 25.1 & 32.0 \\
\cmidrule(r){1-8}
\rowcolor{oursblue} \textbf{NEST (ours)} & \textbf{62.4} & \textbf{43.5} & \textbf{47.3} & \textbf{31.0} & \textbf{41.9} & \textbf{36.4} & \textbf{37.9} \\
\midrule
\multicolumn{8}{l}{\textbf{S-OWODB}} \\
OW-DETR~\cite{gupta2022owdetr} & 71.5 & 5.7 & 43.8 & 6.2 & 38.5 & 6.9 & 33.1 \\
PROB~\cite{zohar2023prob} & 73.4 & 17.6 & 50.4 & 22.3 & 42.0 & 24.8 & 39.9 \\
CAT~\cite{ma2023cat} & \underline{74.2} & 24.0 & 50.7 & 23.0 & 45.0 & 24.6 & 42.8 \\
Hyp-OW~\cite{doan2024hypow} & 72.7 & 23.9 & 50.6 & 23.3 & 46.2 & 25.4 & 44.8 \\
OrthogonalDet~\cite{sun2024exploring} & 71.6 & \underline{24.6} & 51.3 & 27.9 & \textbf{48.8} & \underline{31.9} & \textbf{46.2} \\
Decoupled PROB~\cite{inoue2025decoupled} & 73.7 & 17.4 & \textbf{55.5} & 24.4 & \underline{48.4} & 23.5 & 44.6 \\
PROB+OWOBJ~\cite{zhang2025owobj} & \textbf{76.2} & 22.3 & \underline{54.8} & \underline{28.7} & 46.8 & 30.9 & 43.2 \\
\cmidrule(r){1-8}
\rowcolor{oursblue} \textbf{NEST (ours)} & 74.0 & \textbf{30.8} & 52.2 & \textbf{29.8} & \textbf{48.8} & \textbf{37.3} & \underline{45.5} \\
\midrule
\multicolumn{8}{l}{\textbf{R-OWODB}} \\
RandBox~\cite{wang2023randbox} & \underline{53.3} & 10.5 & 42.2 & 9.7 & \underline{42.4} & 12.8 & 43.6 \\
OrthogonalDet~\cite{sun2024exploring} & \textbf{53.9} & \underline{14.4} & \underline{43.5} & \underline{13.2} & 41.6 & \underline{22.8} & \textbf{45.6} \\
\cmidrule(r){1-8}
\rowcolor{oursblue} \textbf{NEST (ours)} & 52.6 & \textbf{42.5} & \textbf{43.6} & \textbf{33.1} & \textbf{43.8} & \textbf{40.9} & \underline{44.5} \\
\bottomrule
\end{tabular}%
}
\caption{\textbf{Open-world object detection on M-OWODB, S-OWODB and R-OWODB.} Known-class mAP@50 and unknown recall (U-Rec) in \%. NEST achieves the highest U-Rec in every task while remaining competitive in mAP. Bold: best, underline: 2nd best.}
\label{tab:main}
\end{table}

\textbf{Incremental object detection.}\label{sec:iod}
Following prior work~\cite{joseph2021towards,gupta2022owdetr,zohar2023prob,sun2024exploring}, we further evaluate our detector on the standard PASCAL VOC 2007~\cite{everingham2010pascal} benchmark for incremental object detection.
The detector first learns 10, 15 or 19 classes and subsequently learns the remaining 10, 5 or 1 classes, respectively.
We compare with open-world detectors that report this benchmark, using the final \gls{map} over all 20 classes after fine-tuning.
Our detector achieves the highest \gls{map} in the 10+10 and 15+5 settings and is second only to OrthogonalDet in the 19+1 setting (\cref{tab:iod}).
Per-class results are in Supp.~\cref{sec:supp-quant}.

\begin{table}[htbp]
\centering
\footnotesize
\renewcommand{\arraystretch}{0.92}
\setlength{\aboverulesep}{0.25ex}\setlength{\belowrulesep}{0.35ex}
\begin{tabular}{lccc}
\toprule
 & \multicolumn{3}{c}{mAP$\uparrow$} \\
\cmidrule(l){2-4}
Detector & 10+10 & 15+5 & 19+1 \\
\midrule
ORE~\cite{joseph2021towards} & 64.6 & 68.5 & 68.9 \\
OW-DETR~\cite{gupta2022owdetr} & 65.7 & 69.4 & 70.2 \\
PROB~\cite{zohar2023prob} & 66.5 & 70.1 & 72.6 \\
CAT~\cite{ma2023cat} & 67.7 & 72.2 & 73.8 \\
OrthogonalDet~\cite{sun2024exploring} & \underline{72.3} & \underline{74.7} & \textbf{75.6} \\
Decoupled PROB~\cite{inoue2025decoupled} & 67.1 & 70.5 & 73.2 \\
\cmidrule(r){1-4}
\rowcolor{oursblue} \textbf{NEST (ours)}  & \textbf{72.7} & \textbf{75.1} & \underline{74.3} \\
\bottomrule
\end{tabular}
\caption{\textbf{Incremental object detection on PASCAL VOC 2007.} mAP@50 over all 20 classes after fine-tuning, in \%. NEST achieves the highest mAP in the 10+10 and 15+5 settings and the second highest in 19+1. Bold: best, underline: 2nd best.}
\label{tab:iod}
\end{table}

\subsection{Nesting Unknown Objects}
\label{sec:categorization}

We evaluate how well a categorizer assigns unknown objects to their target groups.
A categorizer can only be tested on the unknown objects that its detector finds, so the detector decides what the test covers.
Our detector finds the most unknown objects (\cref{tab:main}), about two to four times as many as RandBox or OrthogonalDet, so we evaluate all categorizers on its R-OWODB detections for the most complete test.
The categorizer assigns a group to the query feature of each prediction labeled \emph{unknown}, and this group is compared with the target $a_t(u)$ of the unknown test object that the prediction matches one-to-one at an \gls{iou} of at least 0.5.

\Gls{owod} methods label unknown objects only as \emph{unknown}, so we adapt the four classification methods of \cref{sec:related_hier} that assign unknown classes to groups, HND~\cite{lee2018hierarchical}, ProHOC~\cite{wallin2025prohoc}, HSC~\cite{linderman2023finegrain} and DARTS~\cite{deng2012hedging}.
Like our categorizer, each learns from the frozen query features of annotated known objects with the same taxonomy, and all five are evaluated on the same matched objects.
We report the mean over three training seeds and give the standard deviations in Supp.~\cref{sec:supp-quant}.
Node-exact measures how often a matched object is assigned to its target group, $\hat g=a_t(u)$.
We also report the \gls{auc} to assess the trade-off between correct assignments to coarser and more specific groups as their relative score bias varies, for Tasks~1--2 only, since at Task~3 no unknown class has the root as its target.
\Gls{e2e} is the share of all unknown test objects that are matched and assigned to their target group, $\hat g=a_t(u)$, so it measures uncovering and nesting together.

\textbf{Results.}
Our categorizer achieves the highest Node-exact, \gls{auc} and \gls{e2e} at every task (\cref{tab:random}).
The task sequence changes what this requires.
At Task~1, about half of the unknown classes have the root \emph{object} as their target, so answering \emph{object} is often correct and all categorizers reach a Node-exact of 56\% to 65\%.
From Task~2 on, most unknown classes have a supercategory as their target, and Node-exact drops for every categorizer.
At Task~3, where answering \emph{object} is never correct, the categorizers differ most.
ProHOC still answers \emph{object} for most matched objects and reaches only 17.8\%, while DARTS rarely answers \emph{object} but often chooses the wrong supercategory.
Our categorizer has its largest lead here, with 49.5\% against 45.4\% for HSC, so it chooses the supercategory more accurately than any baseline.
At Tasks~1 and~2, where the target can be either the root or a supercategory, its highest \gls{auc} also shows that it judges best how specific an answer should be.

Our categorizer needs only one feature vector per detection, making it detector-agnostic.
In Supp.~\cref{sec:supp-quant}, when we also apply it to the features of RandBox or OrthogonalDet, it keeps the highest Node-exact in five of the six cases.

\begin{table}[htbp]
\centering
\footnotesize
\setlength{\tabcolsep}{2.32pt}
\renewcommand{\arraystretch}{1.04}
\begin{tabular}{lcccccccc}
\toprule
 & \multicolumn{3}{c}{Task 1} & \multicolumn{3}{c}{Task 2} & \multicolumn{2}{c}{Task 3} \\
\cmidrule(lr){2-4}\cmidrule(lr){5-7}\cmidrule(l){8-9}
Categorizer & NE$\uparrow$ & AUC$\uparrow$ & E2E$\uparrow$ & NE$\uparrow$ & AUC$\uparrow$ & E2E$\uparrow$ & NE$\uparrow$ & E2E$\uparrow$ \\
\midrule
HND~\cite{lee2018hierarchical} & 56.35 & 58.58 & 27.61 & \underline{36.52} & \underline{30.71} & 15.49 & 33.10 & 17.76 \\
ProHOC~\cite{wallin2025prohoc} & 58.91 & 56.54 & 28.95 & 27.28 & 25.18 & 9.84 & 17.82 & 9.92 \\
HSC~\cite{linderman2023finegrain} & \underline{64.59} & \underline{59.26} & \underline{32.20} & 34.58 & 28.11 & \underline{16.26} & \underline{45.43} & \underline{24.40} \\
DARTS~\cite{deng2012hedging} & 62.12 & 54.20 & 30.93 & 29.68 & 21.62 & 13.79 & 38.22 & 21.44 \\
\cmidrule(r){1-9}
\rowcolor{oursblue} \textbf{NEST (ours)} & \textbf{64.98} & \textbf{59.49} & \textbf{32.22} & \textbf{36.89} & \textbf{31.83} & \textbf{17.39} & \textbf{49.51} & \textbf{26.65} \\
\bottomrule
\end{tabular}
\caption{\textbf{Unknown categorization on R-OWODB with the features of our detector.} Node-exact (NE), AUC and end-to-end recall (E2E) in \%, averaged over three seeds. NEST achieves the highest NE, AUC and E2E at every task. Bold: best, underline: 2nd best.}
\label{tab:random}
\end{table}

\subsection{Qualitative Analysis}
\label{sec:qualitative}

\begin{figure*}[t]
  \centering
  \includegraphics[width=\textwidth]{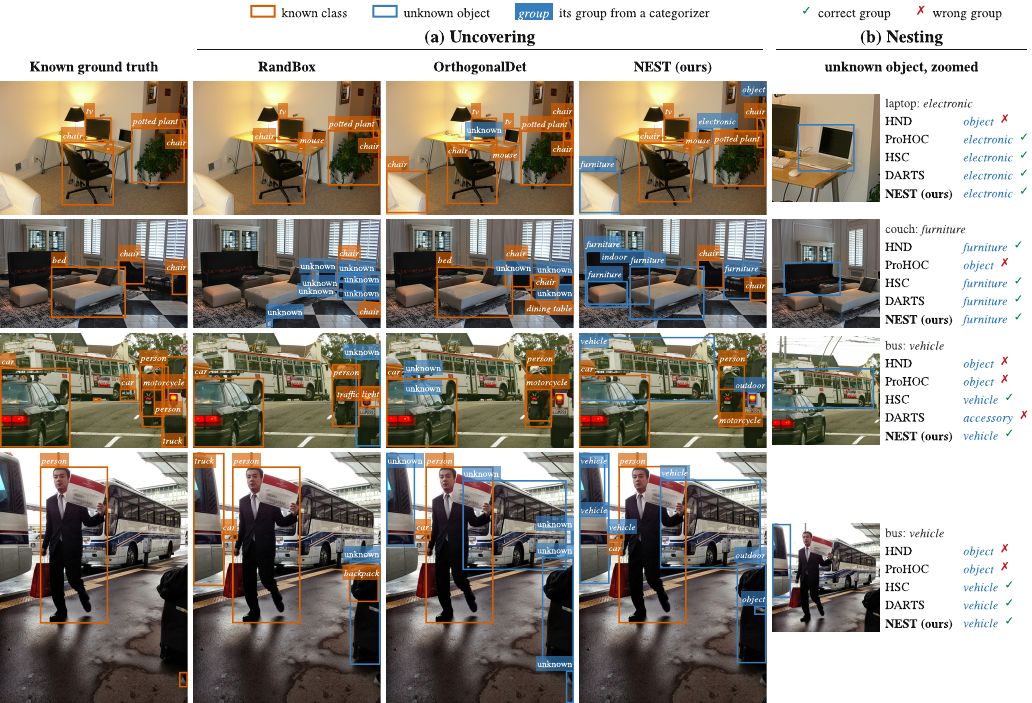}
  \caption{\textbf{Qualitative results.} The first column shows the annotations of the 60 known classes at R-OWODB Task~3. (a)~Uncovering. NEST finds unknown objects that RandBox and OrthogonalDet do not show, such as the laptop and the couches in the first two rows, and also names their group. (b)~Nesting on the same detections from our detector. Each row zooms in on one unknown object, labeled with its class and target group. Where HND and ProHOC fall back to the root \emph{object}, which is never the target at this task, NEST names the target group, such as \emph{electronic} for the laptop and \emph{vehicle} for the buses.}
  \label{fig:qualitative}
\end{figure*}

\Cref{fig:qualitative} shows results on R-OWODB Task~3 after fine-tuning, when 60 classes are known and the other 20 are unknown.
Part~(a) compares the three R-OWODB detectors of \cref{tab:main}, where RandBox and OrthogonalDet label unknown objects only as \emph{unknown} and NEST also names their group.
The detectors score detections on different scales, so we set each detector's display thresholds to show the same average numbers of known and unknown detections per image over all 4167 Task~3 test images.
Known-class detections look similar across the three detectors, consistent with their Task~3 \gls{map} of 41.6 to 43.8 in \cref{tab:main}.
On unknown objects, our NEST detector finds the laptop and the couch in the first row, the two couches in the second and the bus in the third, none of which RandBox or OrthogonalDet shows.
In the last row, OrthogonalDet and our detector both find the two buses, which RandBox misses.

Part~(b) zooms in on one unknown object per image and compares the groups that the categorizers of \cref{tab:random} assign to the same detection from our detector.
HND and ProHOC each answer \emph{object} for three of the four objects, although no unknown class at Task~3 has the root as its target.
DARTS never answers \emph{object} but assigns the bus in the third row to \emph{accessory}.
HSC and our categorizer assign all four to their target groups, in line with their leading Task~3 Node-exact in \cref{tab:random}.
More qualitative results are in Supp.~\cref{sec:supp-qual}.

\section{Ablation Studies}
\label{sec:ablation}

We ablate the parts of NEST that uncover unknown objects (\cref{sec:abl-detector}) and those that nest them in groups (\cref{sec:abl-categorizer}).

\subsection{Detector Components}
\label{sec:abl-detector}

\begin{figure}[t]
  \centering
  \includegraphics[width=\linewidth]{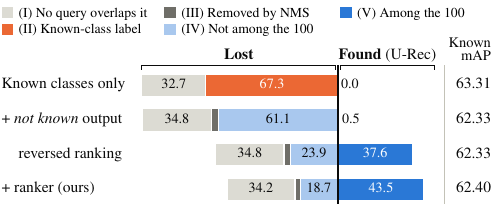}
  \caption{\textbf{Where the unknown objects end up.} Each row splits the 15704 unknown test objects of M-OWODB Task~1 by where one detector loses or finds them, in \%. With known classes only, every unknown object that a query reaches gets a known-class label (II). The \emph{not known} output labels most of them \emph{unknown}, but their detections fall outside the 100 that each image keeps (IV). Ranking the same detections by one minus their \emph{not known} probability already finds 37.6\%, and our ranker finds 43.5\%.}
  \label{fig:abl-where}
\end{figure}

To test how each part of our detector helps it uncover unknown objects while keeping its known \gls{map}, we add the parts of \cref{sec:discovery} one at a time and compare the three resulting variants, shown as rows in \cref{fig:abl-where}.
The first has a class head with outputs for the 20 known classes only.
The second adds the \emph{not known} output.
The third adds the ranker, which completes our detector.

An unknown object can end up in only five places, which follow the order of inference.
For each detector variant, we track which place each of the 15704 unknown objects of the M-OWODB Task~1 test set ends up in.
(I)~No query overlaps it with an \gls{iou} of at least 0.5.
(II)~Every query on it is labeled with a known class.
(III)~\Gls{nms} removes all its detections labeled \emph{unknown}.
(IV)~None of these detections is among the 100 that the image keeps.
(V)~One of the 100 overlaps it and is labeled \emph{unknown}, so the object is found.
\Gls{urec} is the share of objects in this last place.

Without an output for unknown objects, all 67.3\% of the unknown objects that a query reaches end up in place (II).
The detector has no other label to give them, so each one is labeled with whichever known class scores highest.
For example, trucks are labeled \emph{car} in 66\% of cases.
The \emph{not known} output gives these objects a label of their own and shrinks place (II) to 0.4\%.
Most of them move to place (IV) instead (61.1\%).
Their detections labeled \emph{unknown} are outranked by background and miss the 100 that each image keeps, so only 0.52\% of the unknown objects are found.
A likely reason is that the detector is sure that background belongs to no known class, while an unknown object often looks like a known class, as a zebra looks like a horse.
The detector therefore gives background the higher \emph{not known} probability, 0.84 against 0.69 for unknown objects at the median.
A natural question is whether simply reversing this ranking helps, and it does.
Ranked by one minus their \emph{not known} probability, the same detections find 37.6\% of the unknown objects (indented row of \cref{fig:abl-where}).
When our ranker chooses the 100 instead, 43.5\% are found and place (IV) shrinks to 18.7\%, while known \gls{map} stays within one point of the known-only detector (62.40 against 63.31).

\subsection{Categorizer Components}
\label{sec:abl-categorizer}

We ablate the cut loss and the widening $\Delta_g$ of the categorizer of \cref{sec:family_geometry} on R-OWODB (\cref{tab:abl-categorizer}).

Without the cut loss, the categorizer learns only from the known-class loss, which asks each known object to score its own class above all other classes and groups.
A group thus appears in training only as a rival that must lose, and the groups lose most easily by spreading their Gaussians, at Task~1 to ten times the variance of the known classes they contain, against 3.6 times in the full model.
A spread Gaussian gives a low score everywhere, while the root's mixture also holds the known classes directly under it, whose Gaussians stay narrower.
At Task~1, with seven such classes, the categorizer therefore answers \emph{object} for 99.5\% of the objects whose target is a group.
At Tasks~2 and~3, with three and one such classes, it answers \emph{object} for only 1.7\% and 0.4\% of them, but piles them onto single groups, sending 26\% and 29\% to \emph{furniture} and \emph{indoor}, which are the targets of only 6\% and 10\%.

\begin{table}[htbp]
\centering
\footnotesize
\setlength{\tabcolsep}{4.2pt}
\renewcommand{\arraystretch}{1.04}
\begin{tabular}{ccccccc}
\toprule
 & & \multicolumn{2}{c}{Task 1} & \multicolumn{2}{c}{Task 2} & Task 3 \\
\cmidrule(lr){3-4}\cmidrule(lr){5-6}\cmidrule(l){7-7}
Cut loss & $\Delta_g$ & NE$\uparrow$ & AUC$\uparrow$ & NE$\uparrow$ & AUC$\uparrow$ & NE$\uparrow$ \\
\midrule
-- & \checkmark & 48.45 & 14.43 & 30.11 & 25.68 & 44.07 \\
\checkmark & -- & 64.20 & 58.60 & 31.57 & 26.49 & 37.50 \\
\checkmark & \checkmark & \textbf{64.98} & \textbf{59.49} & \textbf{36.89} & \textbf{31.83} & \textbf{49.51} \\
\bottomrule
\end{tabular}
\caption{\textbf{Categorizer ablation on R-OWODB with our detector.} Node-exact (NE) and AUC in \%, averaged over three seeds, with standard deviations of at most 0.67 except 2.11 for the AUC without the cut loss at Task~1. Removing either the cut loss or the increment $\Delta_g$ lowers every score.}
\label{tab:abl-categorizer}
\end{table}

Without the increments $\Delta_g$, each group is made of its members' own narrow Gaussians and scores high only right next to a known member.
The root, however, still covers each group with one wide Gaussian spanning all its members (\cref{eq:recursive_moments}).
A new member that is not right next to a known member therefore goes to \emph{object}, as do 49.8\% and 29.8\% of the objects whose target is a group at Tasks~2 and~3, against 9.8\% and 1.9\% in the full model, and Node-exact falls by 5.32 and 12.01 points.
At Task~1, groups hold at most four members instead of up to seven and eight, so the root's Gaussians cover less extra space, and Node-exact drops by only 0.78.

\section{Conclusion}
\label{sec:conclusion}

Open-world detectors find too few unknown objects and say nothing about what they are.
NEST addresses both problems with two components: a \emph{class-agnostic ranker} that uncovers unknown objects and a \emph{detector-agnostic categorizer} that nests them in the groups of a class taxonomy.
Both learn from the known classes alone.
For uncovering, we show that most unknown objects are not missed by the detector but discarded in its final selection, because the \emph{not known} label mixes them with background.
By judging how well each prediction fits an object, the ranker recovers most of them and gives the highest \gls{urec} in every task of three benchmarks.
For nesting, the categorizer builds each group as a region in feature space that contains its known classes and learns how much room to leave for unseen members.
On our detector, it assigns unknown objects to their target groups more often than four hierarchical classification methods at every task.
With NEST, the unknown objects that a detector already localizes no longer have to be discarded or left only as \emph{unknown}.

{
    \small
    \bibliographystyle{ieeenat_fullname}
    \bibliography{main}
}

\clearpage
\makesupptitle
\glsresetall

\setcounter{section}{0}
\renewcommand{\thesection}{\Alph{section}}
\renewcommand{\theHsection}{supp.\Alph{section}}
\counterwithin{table}{section}
\counterwithin{figure}{section}
\renewcommand{\thetable}{\thesection\arabic{table}}
\renewcommand{\thefigure}{\thesection\arabic{figure}}
\renewcommand{\theHtable}{supp.\thetable}
\renewcommand{\theHfigure}{supp.\thefigure}

\section{Implementation Details}
\label{sec:supp-impl}

\Cref{tab:supp-settings} lists the training settings of our detector and categorizer, and the paragraphs after it explain how we choose the logit adjustment $\tau$ and the ranker threshold $\eta$ on held-out training images.

\begin{table}[htbp]
\centering
\footnotesize
\begin{tabular}{@{}lllll@{}}
\toprule
\multicolumn{5}{@{}l}{\textbf{Detector}} \\
Backbone & \multicolumn{4}{l}{ResNet-50 with FPN, pre-trained on ImageNet} \\
Random boxes & \multicolumn{4}{l}{500 in training, $20\times500$ at test} \\
Stages & \multicolumn{4}{l}{6} \\
Query feature size & \multicolumn{4}{l}{256} \\
NMS IoU threshold & \multicolumn{4}{l}{0.6} \\
Final predictions $k$ & \multicolumn{4}{l}{100 per image} \\
Ranker threshold $\eta$ & \multicolumn{4}{l}{0.3} \\
Logit adjustment $\tau$ & \multicolumn{4}{l}{0.5} \\
Classification loss & \multicolumn{4}{l}{Sigmoid focal~\cite{lin2017focal}} \\
\multicolumn{5}{@{}l}{\textit{Loss weights}} \\
\quad Classification & \multicolumn{4}{l}{2, with the \emph{not known} term $\times0.1$} \\
\quad L1 & \multicolumn{4}{l}{5} \\
\quad GIoU & \multicolumn{4}{l}{2} \\
\quad Ranker & \multicolumn{4}{l}{4} \\
Optimizer & \multicolumn{4}{l}{AdamW} \\
Learning rate & \multicolumn{4}{l}{$2.5\times10^{-5}$} \\
Weight decay & \multicolumn{4}{l}{$10^{-4}$} \\
Batch size & \multicolumn{4}{l}{12 images} \\
\textit{Iterations} & & \multicolumn{2}{c}{\textit{Later tasks}} & \\
\cmidrule(lr){3-4}
 & \textit{Task 1} & \textit{Training} & \textit{Fine-tuning} & \textit{Total} \\
\quad M-OWODB & 20k & $3\times15$k & $3\times15$k & 110k \\
\quad S-OWODB & 100k & $3\times15$k & $3\times15$k & 190k \\
\quad R-OWODB & 40k & $3\times15$k & $3\times15$k & 130k \\
\quad VOC IOD & 18k & $1\times2$k & $1\times3$k & 23k \\
\midrule
\multicolumn{5}{@{}l}{\textbf{Categorizer}} \\
Projection size $d$ & \multicolumn{4}{l}{32} \\
Optimizer & \multicolumn{4}{l}{AdamW} \\
Learning rate & \multicolumn{4}{l}{$5\times10^{-4}$} \\
Weight decay & \multicolumn{4}{l}{$10^{-4}$} \\
Epochs & \multicolumn{4}{l}{at most 50} \\
\bottomrule
\end{tabular}
\caption{\textbf{Training settings.}}
\label{tab:supp-settings}
\end{table}

\textbf{Choosing $\tau$.}
Across benchmarks and tasks, 95\% to 99\% of the training queries receive the \emph{not known} target.
This imbalance inflates the \emph{not known} output.
To mitigate this, the logit adjustment of \cref{sec:discovery} lowers the \emph{not known} logit by $\tau\log(\pi/(1-\pi))$, where $\pi$ is this fraction, and $\tau$ controls the strength of this correction.

Because $\tau$ affects both known \gls{map} and \gls{urec}, it should ideally be chosen on a validation set in which both known and unknown objects are annotated.
However, the \gls{owod} benchmarks have no such validation set, and choosing $\tau$ on the test set would tune it to the test data.
A natural alternative is to hold out part of the training images, but this does not solve the problem.
The M-OWODB Task~1 training images come from PASCAL VOC and are annotated only for its 20 classes, which are all known in Task~1.
A random 10\% of them would therefore contain no labeled unknown objects to measure \gls{urec} on.

To obtain labeled unknown objects, we turn some known classes into unknown ones.
We hide 5 of the 20 known classes, chosen at random, and train our detector on the other 15 classes with a random 90\% of the Task~1 training images and the same settings as for M-OWODB Task~1.
Objects of the hidden classes thus appear in these training images without labels, as unknown objects do in the benchmarks.
The remaining 10\% of the images serve as the validation set.
They keep their annotations of the hidden classes, so \gls{urec} is measured on the 700 objects of these classes and known \gls{map} on the 15 known classes.

\begin{figure}[htbp]
  \centering
  \includegraphics{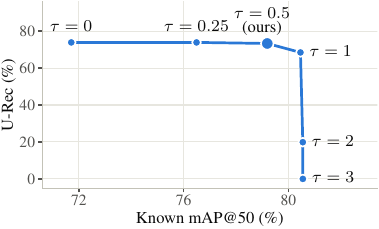}
  \caption{\textbf{Known mAP@50 and U-Rec of our detector as $\tau$ varies} on held-out M-OWODB Task~1 training images.}
  \label{fig:supp-tau}
\end{figure}

For each $\tau$ in \cref{fig:supp-tau}, we measure how far known \gls{map} and \gls{urec} fall below their best values and choose the $\tau$ whose larger drop is smallest.
This rule selects $\tau=0.5$, at the knee of the curve, where known \gls{map} stays within 1.4 points and \gls{urec} within 0.6 points of their best values.
Without the adjustment ($\tau=0$), known \gls{map} is 7.49 points lower than at $\tau=0.5$, for only 0.57 more \gls{urec}.
Larger values of $\tau$ close the 1.4-point gap in known \gls{map} but cost \gls{urec}, which drops to 19.86 at $\tau=2$.
We use $\tau=0.5$ for all tasks and benchmarks.

\begin{figure}[htbp]
  \centering
  \includegraphics{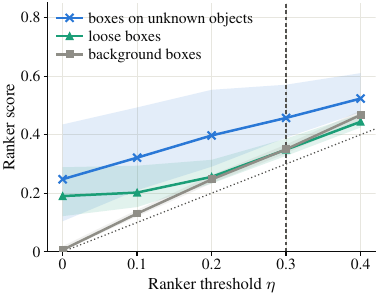}
  \caption{\textbf{Ranker score of three types of predictions as $\eta$ varies} on the validation images of \cref{fig:supp-tau}. Lines show medians and bands the 25th to 75th percentiles. Background boxes score almost alike, so their band is barely visible. Dashed: our $\eta=0.3$. Dotted: score equal to $\eta$, which background boxes follow closely. At a low $\eta$, boxes on unknown objects score barely above loose boxes and compete with them for the 100 kept predictions. At a high $\eta$, background boxes come close to boxes on unknown objects and compete with them instead. At our $\eta=0.3$, boxes on unknown objects stay above both.}
  \label{fig:supp-eta}
\end{figure}

\textbf{Choosing $\eta$.}
The ranker loss of \cref{eq:ranker} covers only the predicted boxes whose \gls{iou} with an annotated object exceeds $\eta$.
The threshold thus decides which predictions the ranker learns from.
We choose it on the same validation set as $\tau$.
In \cref{tab:supp-eta}, $\eta=0.3$ reaches both the highest known \gls{map} and the highest \gls{urec}.
\Cref{fig:supp-eta} shows why.
For each $\eta$, it plots the median ranker score of three types of predictions on the validation images.
\emph{Boxes on unknown objects} overlap an unknown object with an \gls{iou} above 0.5.
\emph{Loose boxes} overlap a known or unknown object only partly, with a best \gls{iou} between 0.1 and 0.5.
\emph{Background boxes} overlap no known or unknown object with an \gls{iou} above 0.1.
Boxes on unknown objects compete with loose boxes and background boxes for the 100 places that each image keeps.
At a low $\eta$, loose boxes are the stronger competitor.
The ranker learns to give boxes with a low \gls{iou} a low score, and since only known objects are annotated during training, boxes on unknown objects count as low-\gls{iou} boxes too and end up barely above loose boxes.
At a high $\eta$, background boxes are the stronger competitor.
With no target below $\eta$, the ranker scores background boxes close to $\eta$, almost as high as boxes on unknown objects.
We use $\eta=0.3$ for all tasks and benchmarks.

\begin{table}[htbp]
\centering
\footnotesize
\begin{tabular}{@{}lccccc@{}}
\toprule
$\eta$ & 0 & 0.1 & 0.2 & 0.3 & 0.4 \\
\midrule
Known mAP@50 & 77.86 & 78.75 & 77.90 & \textbf{79.19} & 79.16 \\
U-Rec & 60.43 & 67.14 & 70.57 & \textbf{73.29} & 70.29 \\
\bottomrule
\end{tabular}
\caption{\textbf{Known mAP@50 and U-Rec of our detector as $\eta$ varies} on the validation images of \cref{fig:supp-tau}, at $\tau=0.5$, in \%. Bold: best.}
\label{tab:supp-eta}
\end{table}

\vspace{-10pt}
\section{Benchmarks and Taxonomy}
\label{sec:supp-splits}

Under the retention rule of \cref{sec:family_geometry}, at least 90\% of the unknown test objects at Tasks~1--3 of M-OWODB and S-OWODB have the root \emph{object} as their target.
On R-OWODB, this share falls from 54.2\% at Task~1 to 17.1\% at Task~2 and 0\% at Task~3 (Root column of \cref{tab:root-targets}).
The difference between the benchmarks comes from the order in which they introduce the classes of each COCO supercategory.
S-OWODB introduces each supercategory within one task, and M-OWODB does the same after the 20 PASCAL VOC classes of Task~1.
The unknown classes of such a supercategory always target the root, since none of its classes is known before that task.
R-OWODB instead shuffles the 80 COCO classes once with a random seed and splits them into four tasks of 20, so most supercategories are spread over several tasks.
In \cref{tab:supp-classes}, which colors each class by its supercategory, each task of S-OWODB and M-OWODB holds a few colors, while each task of R-OWODB mixes most colors.

M-OWODB combines PASCAL VOC~\cite{everingham2010pascal} and COCO images, while S-OWODB and R-OWODB use COCO images alone.
Each task trains on the images that contain at least one of its classes.
\Cref{tab:root-targets} lists the number of training and fine-tuning images in each task.

\definecolor{scperson}{HTML}{B8B8B8}
\definecolor{scvehicle}{HTML}{6C9BD2}
\definecolor{scoutdoor}{HTML}{B5CF6B}
\definecolor{scanimal}{HTML}{F2A65A}
\definecolor{scaccessory}{HTML}{B39DDB}
\definecolor{scsports}{HTML}{63C5C0}
\definecolor{sckitchen}{HTML}{EF8A8A}
\definecolor{scfood}{HTML}{F6D860}
\definecolor{scfurniture}{HTML}{C49A6C}
\definecolor{scelectronic}{HTML}{E6A6D8}
\definecolor{scappliance}{HTML}{8593A3}
\definecolor{scindoor}{HTML}{7FBF7F}
\newcommand{\scchip}[3]{{\setlength{\fboxsep}{0.8pt}\setlength{\fboxrule}{0.5pt}\fcolorbox{#1}{#2!42}{\rule[-0.22em]{0pt}{0.9em}#3}}}
\newcommand{\scwhole}[2]{\scchip{#1}{#1}{#2}}
\newcommand{\scpart}[2]{\scchip{#1!42}{#1}{#2}}
\newcommand{\scsep}{\hspace{1.2pt plus 0.6pt}}
\newcommand{\scgap}{\hspace{4pt plus 1pt}}
\newcommand{\scsq}[1]{{\setlength{\fboxsep}{0pt}\setlength{\fboxrule}{0.4pt}\fcolorbox{#1}{#1!42}{\rule{0pt}{0.6em}\rule{0.6em}{0pt}}}}
\newcommand{\sccnt}[3]{{\setlength{\fboxsep}{0.6pt}\setlength{\fboxrule}{0.5pt}\fcolorbox{#1}{#2!42}{\makebox[0.95em]{#3}}}}
\newcommand{\sccntwhole}[2]{\sccnt{#1}{#1}{#2}}
\newcommand{\sccntpart}[2]{\sccnt{#1!42}{#1}{#2}}

\begin{table}[htbp]
\centering
\footnotesize
\setlength{\tabcolsep}{3.5pt}
\begin{tabular}{@{}lrrrl@{}}
\toprule
 & \multicolumn{2}{c}{Images} & \multicolumn{2}{c}{Unknown test objects} \\
\cmidrule(lr){2-3}\cmidrule(l){4-5}
Task & Train & Fine-tune & Root (\%) & Groups \\
\midrule
\multicolumn{5}{@{}l}{\textbf{M-OWODB}} \\
1 & 16551 & -- & 90.0 & 3\enspace\scsq{scvehicle}\,\scsq{scanimal}\,\scsq{scfurniture} \\
2 & 45520 & 1743 & 96.8 & 1\enspace\scsq{scfurniture} \\
3 & 39402 & 2361 & 94.3 & 1\enspace\scsq{scfurniture} \\
4 & 40260 & 2749 & -- & -- \\
\midrule
\multicolumn{5}{@{}l}{\textbf{S-OWODB}} \\
1 & 89490 & -- & 100.0 & 0 \\
2 & 55870 & 1679 & 100.0 & 0 \\
3 & 39402 & 2345 & 100.0 & 0 \\
4 & 38903 & 2664 & -- & -- \\
\midrule
\multicolumn{5}{@{}l}{\textbf{R-OWODB}} \\
1 & 75650 & -- & 54.2 & 5\enspace\scsq{scvehicle}\,\scsq{scanimal}\,\scsq{scsports}\,\scsq{scfood}\,\scsq{scelectronic} \\
2 & 42861 & 1588 & 17.1 & 9\enspace\scsq{scvehicle}\,\scsq{scoutdoor}\,\scsq{scanimal}\,\scsq{scsports}\,\scsq{sckitchen}\,\scsq{scfood}\,\scsq{scfurniture}\,\scsq{scelectronic}\,\scsq{scindoor} \\
3 & 49263 & 2378 & 0.0 & 9\enspace\scsq{scvehicle}\,\scsq{scanimal}\,\scsq{scaccessory}\,\scsq{scsports}\,\scsq{scfood}\,\scsq{scfurniture}\,\scsq{scelectronic}\,\scsq{scappliance}\,\scsq{scindoor} \\
4 & 42272 & 3151 & -- & -- \\
\bottomrule
\end{tabular}
\caption{\textbf{Per-task statistics.} Train and Fine-tune give the number of training and fine-tuning images. Root is the percentage of unknown test objects whose target is the root. Groups counts the retained groups below the root that are the target of at least one unknown class, with squares in the colors of \cref{tab:supp-classes}. Task~4 has no unknown classes.}
\label{tab:root-targets}
\end{table}

\begin{table*}[htbp]
\centering
\scriptsize
\setlength{\tabcolsep}{2.5pt}
\parbox{\textwidth}{\raggedright \scsq{scperson}\,person\hspace{4pt}\,\scsq{scvehicle}\,vehicle\hspace{4pt}\,\scsq{scoutdoor}\,outdoor\hspace{4pt}\,\scsq{scanimal}\,animal\hspace{4pt}\,\scsq{scaccessory}\,accessory\hspace{4pt}\,\scsq{scsports}\,sports\hspace{4pt}\,\scsq{sckitchen}\,kitchen\hspace{4pt}\,\scsq{scfood}\,food\hspace{4pt}\,\scsq{scfurniture}\,furniture\hspace{4pt}\,\scsq{scelectronic}\,electronic\hspace{4pt}\,\scsq{scappliance}\,appliance\hspace{4pt}\,\scsq{scindoor}\,indoor\hspace{4pt}}\par\vspace{4pt}
\begin{tabular}{@{}>{\raggedright\arraybackslash}p{44pt}*{4}{>{\raggedright\arraybackslash\linespread{1.3}\selectfont}p{107pt}}@{}}
\toprule
 & Task 1 & Task 2 & Task 3 & Task 4 \\
\midrule
\textbf{M-OWODB} & \scwhole{scperson}{person}\scgap\scpart{scvehicle}{bicycle}\scsep\scpart{scvehicle}{car}\scsep\scpart{scvehicle}{motorcycle}\scsep\scpart{scvehicle}{airplane}\scsep\scpart{scvehicle}{bus}\scsep\scpart{scvehicle}{train}\scsep\scpart{scvehicle}{boat}\scgap\scpart{scanimal}{bird}\scsep\scpart{scanimal}{cat}\scsep\scpart{scanimal}{dog}\scsep\scpart{scanimal}{horse}\scsep\scpart{scanimal}{sheep}\scsep\scpart{scanimal}{cow}\scgap\scpart{sckitchen}{bottle}\scgap\scpart{scfurniture}{chair}\scsep\scpart{scfurniture}{couch}\scsep\scpart{scfurniture}{potted plant}\scsep\scpart{scfurniture}{dining table}\scgap\scpart{scelectronic}{tv} & \scpart{scvehicle}{truck}\scgap\scwhole{scoutdoor}{traffic light}\scsep\scwhole{scoutdoor}{fire hydrant}\scsep\scwhole{scoutdoor}{stop sign}\scsep\scwhole{scoutdoor}{parking meter}\scsep\scwhole{scoutdoor}{bench}\scgap\scpart{scanimal}{elephant}\scsep\scpart{scanimal}{bear}\scsep\scpart{scanimal}{zebra}\scsep\scpart{scanimal}{giraffe}\scgap\scwhole{scaccessory}{backpack}\scsep\scwhole{scaccessory}{umbrella}\scsep\scwhole{scaccessory}{handbag}\scsep\scwhole{scaccessory}{tie}\scsep\scwhole{scaccessory}{suitcase}\scgap\scwhole{scappliance}{microwave}\scsep\scwhole{scappliance}{oven}\scsep\scwhole{scappliance}{toaster}\scsep\scwhole{scappliance}{sink}\scsep\scwhole{scappliance}{refrigerator} & \scwhole{scsports}{frisbee}\scsep\scwhole{scsports}{skis}\scsep\scwhole{scsports}{snowboard}\scsep\scwhole{scsports}{sports ball}\scsep\scwhole{scsports}{kite}\scsep\scwhole{scsports}{baseball bat}\scsep\scwhole{scsports}{baseball glove}\scsep\scwhole{scsports}{skateboard}\scsep\scwhole{scsports}{surfboard}\scsep\scwhole{scsports}{tennis racket}\scgap\scwhole{scfood}{banana}\scsep\scwhole{scfood}{apple}\scsep\scwhole{scfood}{sandwich}\scsep\scwhole{scfood}{orange}\scsep\scwhole{scfood}{broccoli}\scsep\scwhole{scfood}{carrot}\scsep\scwhole{scfood}{hot dog}\scsep\scwhole{scfood}{pizza}\scsep\scwhole{scfood}{donut}\scsep\scwhole{scfood}{cake} & \scpart{sckitchen}{wine glass}\scsep\scpart{sckitchen}{cup}\scsep\scpart{sckitchen}{fork}\scsep\scpart{sckitchen}{knife}\scsep\scpart{sckitchen}{spoon}\scsep\scpart{sckitchen}{bowl}\scgap\scpart{scfurniture}{bed}\scsep\scpart{scfurniture}{toilet}\scgap\scpart{scelectronic}{laptop}\scsep\scpart{scelectronic}{mouse}\scsep\scpart{scelectronic}{remote}\scsep\scpart{scelectronic}{keyboard}\scsep\scpart{scelectronic}{cell phone}\scgap\scwhole{scindoor}{book}\scsep\scwhole{scindoor}{clock}\scsep\scwhole{scindoor}{vase}\scsep\scwhole{scindoor}{scissors}\scsep\scwhole{scindoor}{teddy bear}\scsep\scwhole{scindoor}{hair drier}\scsep\scwhole{scindoor}{toothbrush} \\
\midrule
\textbf{S-OWODB} & \scwhole{scperson}{person}\scgap\scwhole{scvehicle}{bicycle}\scsep\scwhole{scvehicle}{car}\scsep\scwhole{scvehicle}{motorcycle}\scsep\scwhole{scvehicle}{airplane}\scsep\scwhole{scvehicle}{bus}\scsep\scwhole{scvehicle}{train}\scsep\scwhole{scvehicle}{truck}\scsep\scwhole{scvehicle}{boat}\scgap\scwhole{scanimal}{bird}\scsep\scwhole{scanimal}{cat}\scsep\scwhole{scanimal}{dog}\scsep\scwhole{scanimal}{horse}\scsep\scwhole{scanimal}{sheep}\scsep\scwhole{scanimal}{cow}\scsep\scwhole{scanimal}{elephant}\scsep\scwhole{scanimal}{bear}\scsep\scwhole{scanimal}{zebra}\scsep\scwhole{scanimal}{giraffe} & \scwhole{scoutdoor}{traffic light}\scsep\scwhole{scoutdoor}{fire hydrant}\scsep\scwhole{scoutdoor}{stop sign}\scsep\scwhole{scoutdoor}{parking meter}\scsep\scwhole{scoutdoor}{bench}\scgap\scwhole{scaccessory}{backpack}\scsep\scwhole{scaccessory}{umbrella}\scsep\scwhole{scaccessory}{handbag}\scsep\scwhole{scaccessory}{tie}\scsep\scwhole{scaccessory}{suitcase}\scgap\scwhole{scfurniture}{chair}\scsep\scwhole{scfurniture}{couch}\scsep\scwhole{scfurniture}{potted plant}\scsep\scwhole{scfurniture}{bed}\scsep\scwhole{scfurniture}{dining table}\scsep\scwhole{scfurniture}{toilet}\scgap\scwhole{scappliance}{microwave}\scsep\scwhole{scappliance}{oven}\scsep\scwhole{scappliance}{toaster}\scsep\scwhole{scappliance}{sink}\scsep\scwhole{scappliance}{refrigerator} & \scwhole{scsports}{frisbee}\scsep\scwhole{scsports}{skis}\scsep\scwhole{scsports}{snowboard}\scsep\scwhole{scsports}{sports ball}\scsep\scwhole{scsports}{kite}\scsep\scwhole{scsports}{baseball bat}\scsep\scwhole{scsports}{baseball glove}\scsep\scwhole{scsports}{skateboard}\scsep\scwhole{scsports}{surfboard}\scsep\scwhole{scsports}{tennis racket}\scgap\scwhole{scfood}{banana}\scsep\scwhole{scfood}{apple}\scsep\scwhole{scfood}{sandwich}\scsep\scwhole{scfood}{orange}\scsep\scwhole{scfood}{broccoli}\scsep\scwhole{scfood}{carrot}\scsep\scwhole{scfood}{hot dog}\scsep\scwhole{scfood}{pizza}\scsep\scwhole{scfood}{donut}\scsep\scwhole{scfood}{cake} & \scwhole{sckitchen}{bottle}\scsep\scwhole{sckitchen}{wine glass}\scsep\scwhole{sckitchen}{cup}\scsep\scwhole{sckitchen}{fork}\scsep\scwhole{sckitchen}{knife}\scsep\scwhole{sckitchen}{spoon}\scsep\scwhole{sckitchen}{bowl}\scgap\scwhole{scelectronic}{tv}\scsep\scwhole{scelectronic}{laptop}\scsep\scwhole{scelectronic}{mouse}\scsep\scwhole{scelectronic}{remote}\scsep\scwhole{scelectronic}{keyboard}\scsep\scwhole{scelectronic}{cell phone}\scgap\scwhole{scindoor}{book}\scsep\scwhole{scindoor}{clock}\scsep\scwhole{scindoor}{vase}\scsep\scwhole{scindoor}{scissors}\scsep\scwhole{scindoor}{teddy bear}\scsep\scwhole{scindoor}{hair drier}\scsep\scwhole{scindoor}{toothbrush} \\
\midrule
\textbf{R-OWODB} & \scwhole{scperson}{person}\scgap\scpart{scvehicle}{bicycle}\scsep\scpart{scvehicle}{train}\scgap\scpart{scoutdoor}{parking meter}\scgap\scpart{scanimal}{dog}\scsep\scpart{scanimal}{giraffe}\scgap\scpart{scaccessory}{umbrella}\scgap\scpart{scsports}{skis}\scsep\scpart{scsports}{snowboard}\scsep\scpart{scsports}{kite}\scgap\scpart{sckitchen}{cup}\scgap\scpart{scfood}{banana}\scsep\scpart{scfood}{sandwich}\scsep\scpart{scfood}{hot dog}\scsep\scpart{scfood}{cake}\scgap\scpart{scfurniture}{chair}\scgap\scpart{scelectronic}{mouse}\scsep\scpart{scelectronic}{cell phone}\scgap\scpart{scappliance}{toaster}\scgap\scpart{scindoor}{clock} & \scpart{scvehicle}{airplane}\scsep\scpart{scvehicle}{boat}\scgap\scpart{scoutdoor}{stop sign}\scsep\scpart{scoutdoor}{bench}\scgap\scpart{scanimal}{bird}\scsep\scpart{scanimal}{cat}\scsep\scpart{scanimal}{sheep}\scsep\scpart{scanimal}{elephant}\scsep\scpart{scanimal}{zebra}\scgap\scpart{scsports}{skateboard}\scgap\scpart{sckitchen}{wine glass}\scsep\scpart{sckitchen}{fork}\scsep\scpart{sckitchen}{knife}\scsep\scpart{sckitchen}{spoon}\scsep\scpart{sckitchen}{bowl}\scgap\scpart{scfood}{carrot}\scgap\scpart{scfurniture}{potted plant}\scsep\scpart{scfurniture}{bed}\scsep\scpart{scfurniture}{dining table}\scgap\scpart{scindoor}{teddy bear} & \scpart{scvehicle}{car}\scsep\scpart{scvehicle}{motorcycle}\scsep\scpart{scvehicle}{truck}\scgap\scpart{scoutdoor}{traffic light}\scsep\scpart{scoutdoor}{fire hydrant}\scgap\scpart{scanimal}{cow}\scgap\scpart{scaccessory}{backpack}\scgap\scpart{scsports}{frisbee}\scsep\scpart{scsports}{sports ball}\scsep\scpart{scsports}{baseball glove}\scsep\scpart{scsports}{surfboard}\scgap\scpart{sckitchen}{bottle}\scgap\scpart{scfood}{broccoli}\scsep\scpart{scfood}{pizza}\scgap\scpart{scelectronic}{tv}\scgap\scpart{scappliance}{microwave}\scsep\scpart{scappliance}{refrigerator}\scgap\scpart{scindoor}{vase}\scsep\scpart{scindoor}{hair drier}\scsep\scpart{scindoor}{toothbrush} & \scpart{scvehicle}{bus}\scgap\scpart{scanimal}{horse}\scsep\scpart{scanimal}{bear}\scgap\scpart{scaccessory}{handbag}\scsep\scpart{scaccessory}{tie}\scsep\scpart{scaccessory}{suitcase}\scgap\scpart{scsports}{baseball bat}\scsep\scpart{scsports}{tennis racket}\scgap\scpart{scfood}{apple}\scsep\scpart{scfood}{orange}\scsep\scpart{scfood}{donut}\scgap\scpart{scfurniture}{couch}\scsep\scpart{scfurniture}{toilet}\scgap\scpart{scelectronic}{laptop}\scsep\scpart{scelectronic}{remote}\scsep\scpart{scelectronic}{keyboard}\scgap\scpart{scappliance}{oven}\scsep\scpart{scappliance}{sink}\scgap\scpart{scindoor}{book}\scsep\scpart{scindoor}{scissors} \\
\bottomrule
\end{tabular}
\caption{\textbf{Classes introduced at each task, colored by COCO supercategory.} M-OWODB and S-OWODB introduce most supercategories within a single task, while R-OWODB spreads almost all of them over several tasks.}
\label{tab:supp-classes}
\end{table*}

\vspace{-10pt}
\section{More Quantitative Results}
\label{sec:supp-quant}

\begin{table*}[htbp]
\centering
\footnotesize
\setlength{\tabcolsep}{4.03pt}
\renewcommand{\arraystretch}{0.92}
\setlength{\aboverulesep}{0.25ex}\setlength{\belowrulesep}{0.35ex}
\begin{tabular}{l*{13}{c}}
\toprule
 & \multicolumn{2}{c}{Task 1} & \multicolumn{4}{c}{Task 2} & \multicolumn{4}{c}{Task 3} & \multicolumn{3}{c}{Task 4} \\
\cmidrule(lr){2-3}\cmidrule(lr){4-7}\cmidrule(lr){8-11}\cmidrule(l){12-14}
Method & Cur.$\uparrow$ & U-Rec$\uparrow$ & Prev.$\uparrow$ & Cur.$\uparrow$ & Both$\uparrow$ & U-Rec$\uparrow$ & Prev.$\uparrow$ & Cur.$\uparrow$ & Both$\uparrow$ & U-Rec$\uparrow$ & Prev.$\uparrow$ & Cur.$\uparrow$ & Both$\uparrow$ \\
\midrule
\multicolumn{14}{l}{\textbf{M-OWODB}} \\
OW-DETR~\cite{gupta2022owdetr} & 59.2 & 7.5 & 53.6 & 33.5 & 42.9 & 6.2 & 38.3 & 15.8 & 30.8 & 5.7 & 31.4 & 17.1 & 27.8 \\
UC-OWOD~\cite{wu2022ucowod} & 50.7 & 2.4 & 33.1 & 30.5 & 31.8 & 3.4 & 28.8 & 16.3 & 24.6 & 8.7 & 25.6 & 15.9 & 23.1 \\
PROB~\cite{zohar2023prob} & 59.5 & 19.4 & 55.7 & 32.2 & 44.0 & 17.4 & 43.0 & 22.2 & 36.0 & 19.6 & 35.7 & 18.9 & 31.5 \\
CAT~\cite{ma2023cat} & 60.0 & 23.7 & 55.5 & 32.7 & 44.1 & 19.1 & 42.8 & 18.7 & 34.8 & 24.4 & 34.4 & 16.6 & 29.9 \\
RandBox~\cite{wang2023randbox} & \underline{61.8} & 10.6 & -- & -- & 45.3 & 6.3 & -- & -- & 39.4 & 7.8 & -- & -- & \underline{35.4} \\
Hyp-OW~\cite{doan2024hypow} & 59.4 & 23.5 & 51.4 & 37.4 & 44.4 & 20.6 & 42.9 & 24.6 & 36.8 & 26.3 & 37.4 & 22.4 & 33.6 \\
OrthogonalDet~\cite{sun2024exploring} & 61.3 & \underline{24.6} & 55.5 & \textbf{38.5} & \underline{47.0} & \underline{26.3} & \underline{46.7} & \underline{30.6} & \underline{41.3} & \underline{29.1} & \textbf{42.4} & \underline{24.3} & \textbf{37.9} \\
Decoupled PROB~\cite{inoue2025decoupled} & 59.8 & 20.3 & \underline{56.4} & 36.7 & 46.6 & 18.4 & 44.6 & 26.1 & 38.5 & 20.3 & 37.8 & 22.1 & 33.8 \\
PROB+OWOBJ~\cite{zhang2025owobj} & 61.4 & 23.6 & \textbf{58.4} & 34.4 & 45.7 & 23.8 & 44.8 & 27.8 & 38.8 & 25.1 & 36.4 & 20.7 & 32.0 \\
\cmidrule(r){1-14}
\rowcolor{oursblue} \textbf{NEST (ours)} & \textbf{62.4} & \textbf{43.5} & \underline{56.4} & \underline{38.2} & \textbf{47.3} & \textbf{31.0} & \textbf{47.3} & \textbf{31.3} & \textbf{41.9} & \textbf{36.4} & \underline{42.3} & \textbf{25.0} & \textbf{37.9} \\
\midrule
\multicolumn{14}{l}{\textbf{S-OWODB}} \\
OW-DETR~\cite{gupta2022owdetr} & 71.5 & 5.7 & 62.8 & 27.5 & 43.8 & 6.2 & 45.2 & 24.9 & 38.5 & 6.9 & 38.2 & 28.1 & 33.1 \\
PROB~\cite{zohar2023prob} & 73.4 & 17.6 & 66.3 & 36.0 & 50.4 & 22.3 & 47.8 & 30.4 & 42.0 & 24.8 & 42.6 & 31.7 & 39.9 \\
CAT~\cite{ma2023cat} & \underline{74.2} & 24.0 & \underline{67.6} & 35.5 & 50.7 & 23.0 & 51.2 & 32.6 & 45.0 & 24.6 & 45.4 & 35.1 & 42.8 \\
Hyp-OW~\cite{doan2024hypow} & 72.7 & 23.9 & 59.8 & \underline{42.2} & 50.6 & 23.3 & 49.3 & 39.8 & 46.2 & 25.4 & 46.4 & \textbf{40.3} & 44.8 \\
OrthogonalDet~\cite{sun2024exploring} & 71.6 & \underline{24.6} & 64.0 & 39.9 & 51.3 & 27.9 & \underline{52.1} & \underline{42.2} & \textbf{48.8} & \underline{31.9} & \textbf{48.7} & 38.8 & \textbf{46.2} \\
Decoupled PROB~\cite{inoue2025decoupled} & 73.7 & 17.4 & 66.3 & \textbf{45.7} & \textbf{55.5} & 24.4 & 50.9 & \textbf{43.3} & \underline{48.4} & 23.5 & 46.1 & \underline{39.8} & 44.6 \\
PROB+OWOBJ~\cite{zhang2025owobj} & \textbf{76.2} & 22.3 & \textbf{69.8} & 41.0 & \underline{54.8} & \underline{28.7} & 50.6 & 35.7 & 46.8 & 30.9 & 46.7 & 36.9 & 43.2 \\
\cmidrule(r){1-14}
\rowcolor{oursblue} \textbf{NEST (ours)} & 74.0 & \textbf{30.8} & 66.1 & 39.7 & 52.2 & \textbf{29.8} & \textbf{52.7} & 40.8 & \textbf{48.8} & \textbf{37.3} & \underline{47.6} & 39.2 & \underline{45.5} \\
\midrule
\multicolumn{14}{l}{\textbf{R-OWODB}} \\
RandBox~\cite{wang2023randbox} & \underline{53.3} & 10.5 & \underline{43.5} & 40.9 & 42.2 & 9.7 & \underline{42.2} & 42.9 & \underline{42.4} & 12.8 & 43.6 & 43.8 & 43.6 \\
OrthogonalDet~\cite{sun2024exploring} & \textbf{53.9} & \underline{14.4} & \textbf{45.3} & \underline{41.7} & \underline{43.5} & \underline{13.2} & 40.8 & \underline{43.3} & 41.6 & \underline{22.8} & \textbf{45.5} & \textbf{45.9} & \textbf{45.6} \\
\cmidrule(r){1-14}
\rowcolor{oursblue} \textbf{NEST (ours)} & 52.6 & \textbf{42.5} & \textbf{45.3} & \textbf{42.0} & \textbf{43.6} & \textbf{33.1} & \textbf{43.0} & \textbf{45.5} & \textbf{43.8} & \textbf{40.9} & \underline{44.5} & \underline{44.3} & \underline{44.5} \\
\bottomrule
\end{tabular}
\caption{\textbf{Detection breakdowns.} Known-class mAP@50, split into previously known (Prev.), currently known (Cur.) and all known classes (Both), and unknown recall (U-Rec) in \%. Dashes: not reported. Bold: best, underline: 2nd best per benchmark and column, including ties at displayed precision.}
\label{tab:supp-det}
\end{table*}

Besides the experiments in \cref{sec:experiments}, we report further results on \gls{owod}, incremental object detection and unknown object categorization.
\Cref{sec:supp-owod} splits the known-class \gls{map} of \cref{tab:main} into previously and currently known classes, reports the \gls{aose} and compares \gls{urec} when our detector keeps only as many unknown predictions as RandBox or OrthogonalDet.
\Cref{sec:supp-iod} gives the \gls{ap} of each class in the incremental setting of \cref{tab:iod}.
\Cref{sec:supp-cat} evaluates all five categorizers on the features of RandBox and OrthogonalDet in addition to ours, gives their standard deviations over three training runs and tests the cut weighting $\rho_t$ of \cref{sec:family_geometry}.

\subsection{Open-World Object Detection}
\label{sec:supp-owod}

\textbf{Baselines.}
As in \cref{sec:owod}, RandBox~\cite{wang2023randbox} and its extension OrthogonalDet~\cite{sun2024exploring} are our most direct comparisons, since we build on RandBox.
On R-OWODB, we train both with their official implementations on the same class partition and task sequence as our detector, with the schedule of \cref{tab:supp-settings}.
On M-OWODB and S-OWODB, we take all baseline results from the corresponding publications.
For a broader comparison on these two benchmarks, we add the DETR-based methods OW-DETR~\cite{gupta2022owdetr}, PROB~\cite{zohar2023prob}, CAT~\cite{ma2023cat}, Hyp-OW~\cite{doan2024hypow}, Decoupled PROB~\cite{inoue2025decoupled} and PROB+OWOBJ~\cite{zhang2025owobj} and, on M-OWODB, the Faster R-CNN-based UC-OWOD~\cite{wu2022ucowod}.

We leave out two other recent \gls{owod} methods, CROWD~\cite{majee2025crowd} and DEUS~\cite{heo2026deus}, because they keep a varying number of predictions per image.
Predictions that overlap no annotated object do NOT lower \gls{urec}, so \textbf{in the extreme case, labeling a dense grid of boxes as unknown would find almost every unknown object}.
Like most \gls{owod} methods, our detector keeps 100 predictions per image, chosen by its ranker, while the public code of CROWD keeps every prediction that survives \gls{nms} and a score threshold, up to 10000 per image.
DEUS has no public code, but its supplementary material describes the same kind of selection as CROWD.

\textbf{Metrics.}
\Cref{tab:supp-det} splits the known-class \gls{map} of \cref{tab:main} into the previously known classes $\mathcal{K}_{t-1}$ (Prev.) and the classes $\mathcal{N}_t$ introduced at task~$t$ (Cur.).
Prev.\ shows how well a detector keeps the earlier classes, and Cur.\ how well it learns the new ones.
The annotations of unknown objects are sparse in the test images.
A prediction labeled unknown that matches no annotation may therefore still be a real object, and a precision-based measure such as unknown \gls{ap} would count it as an error, which is why \gls{owod} methods report the recall of unknown objects.
The sparse annotations are enough, however, for the \gls{aose}, which counts how often a detector labels an unknown test object with a known class.

\textbf{Results.}
In \cref{tab:supp-det}, our detector has the highest \gls{urec} in all nine tasks, as in \cref{tab:main}.
On R-OWODB, where all three detectors follow the same task sequence and schedule, it recalls 1.8 to 4.0 times as many unknown objects as RandBox or OrthogonalDet.
At the same time, its Prev.\ is the best or second best in eight of the nine tasks from Task~2 on, and its Cur.\ in seven of the twelve tasks.
It thus finds more unknown objects than any baseline while remaining competitive on the known classes.

In \cref{tab:supp-aose}, we compare the \gls{aose} on M-OWODB and R-OWODB.
Our detector has the lowest \gls{aose} in three of the six tasks and the second lowest in the other three.
In those three, PROB+OWOBJ at M-OWODB Tasks~1 and~3 and OrthogonalDet at R-OWODB Task~3 label unknown test objects with a known class less often, but our detector recalls 1.5 to 1.8 times as many unknown objects as they do.

\begin{table}[htbp]
\centering
\footnotesize
\setlength{\tabcolsep}{2.58pt}
\renewcommand{\arraystretch}{1.04}
\resizebox{\columnwidth}{!}{%
\begin{tabular}{lcccccc}
\toprule
 & \multicolumn{2}{c}{Task 1} & \multicolumn{2}{c}{Task 2} & \multicolumn{2}{c}{Task 3} \\
\cmidrule(lr){2-3}\cmidrule(lr){4-5}\cmidrule(l){6-7}
Method & U-Rec$\uparrow$ & A-OSE$\downarrow$ & U-Rec$\uparrow$ & A-OSE$\downarrow$ & U-Rec$\uparrow$ & A-OSE$\downarrow$ \\
\midrule
\multicolumn{7}{l}{\textbf{M-OWODB}} \\
OW-DETR~\cite{gupta2022owdetr} & 7.5 & 10240 & 6.2 & 8441 & 5.7 & 6803 \\
UC-OWOD~\cite{wu2022ucowod} & 2.4 & 9294 & 3.4 & 5602 & 8.7 & 3801 \\
PROB~\cite{zohar2023prob} & 19.4 & 5195 & 17.4 & 6452 & 19.6 & 2641 \\
RandBox~\cite{wang2023randbox} & 10.6 & 4498 & 6.3 & \underline{1880} & 7.8 & 1452 \\
Hyp-OW~\cite{doan2024hypow} & 23.5 & 11275 & 20.6 & 4805 & \underline{26.3} & 3548 \\
PROB+OWOBJ~\cite{zhang2025owobj} & \underline{23.6} & \textbf{3102} & \underline{23.8} & 4032 & 25.1 & \textbf{821} \\
\cmidrule(r){1-7}
\rowcolor{oursblue} \textbf{NEST (ours)} & \textbf{43.5} & \underline{4376} & \textbf{31.0} & \textbf{1868} & \textbf{36.4} & \underline{1263} \\
\midrule
\multicolumn{7}{l}{\textbf{R-OWODB}} \\
RandBox~\cite{wang2023randbox} & 10.5 & 3582 & 9.7 & 2177 & 12.8 & 1592 \\
OrthogonalDet~\cite{sun2024exploring} & \underline{14.4} & \underline{2018} & \underline{13.2} & \underline{1501} & \underline{22.8} & \textbf{689} \\
\cmidrule(r){1-7}
\rowcolor{oursblue} \textbf{NEST (ours)} & \textbf{42.5} & \textbf{1779} & \textbf{33.1} & \textbf{1324} & \textbf{40.9} & \underline{878} \\
\bottomrule
\end{tabular}%
}
\caption{\textbf{Absolute open-set error (A-OSE).} Bold: best, underline: 2nd best.}
\label{tab:supp-aose}
\end{table}

On R-OWODB, every detector keeps at most 100 predictions per image, and our detector spends more of this budget on unknown predictions.
In \cref{tab:supp-budget}, we test whether this alone explains its higher \gls{urec}.
For each baseline, we count its unknown predictions across all test images.
We then sort all unknown predictions of our detector by their ranker scores and keep the same number from the top.
With the same average number of unknown predictions per test image (Unk.), our detector still recalls 1.2 to 1.4 times as many unknown objects as RandBox and 1.4 to 1.6 times as many as OrthogonalDet.
Only part of its gain therefore comes from spending more of the budget on unknown predictions.

\begin{table}[htbp]
\centering
\footnotesize
\setlength{\tabcolsep}{2.2pt}
\renewcommand{\arraystretch}{1.04}
\resizebox{0.87\columnwidth}{!}{%
\begin{tabular}{lcccccc}
\toprule
 & \multicolumn{2}{c}{Task 1} & \multicolumn{2}{c}{Task 2} & \multicolumn{2}{c}{Task 3} \\
\cmidrule(lr){2-3}\cmidrule(lr){4-5}\cmidrule(l){6-7}
Detector & Unk. & U-Rec$\uparrow$ & Unk. & U-Rec$\uparrow$ & Unk. & U-Rec$\uparrow$ \\
\midrule
\multicolumn{7}{l}{\textit{Same number as RandBox}} \\
RandBox~\cite{wang2023randbox} & 5.1 & 10.5 & 7.9 & 9.7 & 8.7 & 12.8 \\
\rowcolor{oursblue} NEST (ours) & 5.1 & \textbf{12.2} & 7.9 & \textbf{12.4} & 8.7 & \textbf{18.4} \\
\midrule
\multicolumn{7}{l}{\textit{Same number as OrthogonalDet}} \\
OrthogonalDet~\cite{sun2024exploring} & 12.1 & 14.4 & 20.8 & 13.2 & 53.1 & 22.8 \\
\rowcolor{oursblue} NEST (ours) & 12.1 & \textbf{20.7} & 20.8 & \textbf{21.0} & 53.1 & \textbf{37.2} \\
\bottomrule
\end{tabular}%
}
\caption{\textbf{U-Rec at the same number of unknown predictions on R-OWODB.} Unk.\ is the average number of unknown predictions per test image. Bold: higher U-Rec within each block.}
\label{tab:supp-budget}
\end{table}

\subsection{Incremental Object Detection}
\label{sec:supp-iod}

\Cref{tab:supp-iod} gives the \gls{ap} of each class for the methods and settings of \cref{tab:iod}.
Its Old column shows how well a detector keeps the classes of the first stage, and its New column how well it learns those of the second.
Our detector is the best or second best in both columns in all three settings.
OrthogonalDet is the only method ahead of it, in New at 15+5 and 19+1, with 71.8 and 74.9 against 67.4 and 72.5, and in Old at 19+1, with 75.7 against 74.5.

\colorlet{iodnew}{black!8}
\begin{table*}[htbp]
\centering
\scriptsize
\setlength{\tabcolsep}{2.3pt}
\renewcommand{\arraystretch}{0.92}
\setlength{\aboverulesep}{0.25ex}\setlength{\belowrulesep}{0.35ex}
\begin{tabular}{l*{23}{c}}
\toprule
\textbf{10+10} & aero & cycle & bird & boat & bottle & bus & car & cat & chair & cow & table & dog & horse & bike & person & plant & sheep & sofa & train & tv & \textbf{Old} & \textbf{New} & \textbf{mAP} \\
\midrule
ORE~\cite{joseph2021towards} & 63.5 & 70.9 & 58.9 & 42.9 & 34.1 & 76.2 & 80.7 & 76.3 & 34.1 & 66.1 & \cellcolor{iodnew}56.1 & \cellcolor{iodnew}70.4 & \cellcolor{iodnew}80.2 & \cellcolor{iodnew}72.3 & \cellcolor{iodnew}81.8 & \cellcolor{iodnew}42.7 & \cellcolor{iodnew}71.6 & \cellcolor{iodnew}68.1 & \cellcolor{iodnew}77.0 & \cellcolor{iodnew}67.7 & 60.4 & 68.8 & 64.6 \\
OW-DETR~\cite{gupta2022owdetr} & 61.8 & 69.1 & 67.8 & 45.8 & 47.3 & 78.3 & 78.4 & 78.6 & 36.2 & 71.5 & \cellcolor{iodnew}57.5 & \cellcolor{iodnew}75.3 & \cellcolor{iodnew}76.2 & \cellcolor{iodnew}77.4 & \cellcolor{iodnew}79.5 & \cellcolor{iodnew}40.1 & \cellcolor{iodnew}66.8 & \cellcolor{iodnew}66.3 & \cellcolor{iodnew}75.6 & \cellcolor{iodnew}64.1 & 63.5 & 67.9 & 65.7 \\
PROB~\cite{zohar2023prob} & 70.4 & 75.4 & 67.3 & 48.1 & 55.9 & 73.5 & 78.5 & 75.4 & 42.8 & 72.2 & \cellcolor{iodnew}64.2 & \cellcolor{iodnew}73.8 & \cellcolor{iodnew}76.0 & \cellcolor{iodnew}74.8 & \cellcolor{iodnew}75.3 & \cellcolor{iodnew}40.2 & \cellcolor{iodnew}66.2 & \cellcolor{iodnew}73.3 & \cellcolor{iodnew}64.4 & \cellcolor{iodnew}64.0 & 66.0 & 67.2 & 66.5 \\
CAT~\cite{ma2023cat} & 76.5 & 75.7 & 67.0 & 51.0 & 62.4 & 73.2 & 82.3 & 83.7 & 42.7 & 64.4 & \cellcolor{iodnew}56.8 & \cellcolor{iodnew}74.1 & \cellcolor{iodnew}75.8 & \cellcolor{iodnew}79.2 & \cellcolor{iodnew}78.1 & \cellcolor{iodnew}39.9 & \cellcolor{iodnew}65.1 & \cellcolor{iodnew}59.6 & \cellcolor{iodnew}78.4 & \cellcolor{iodnew}67.4 & 67.9 & 67.4 & 67.7 \\
OrthogonalDet~\cite{sun2024exploring} & 82.4 & 77.3 & 78.2 & 59.7 & 61.2 & 84.3 & 90.1 & 80.2 & 49.8 & 81.7 & \cellcolor{iodnew}58.2 & \cellcolor{iodnew}74.0 & \cellcolor{iodnew}82.9 & \cellcolor{iodnew}81.0 & \cellcolor{iodnew}81.2 & \cellcolor{iodnew}38.3 & \cellcolor{iodnew}70.8 & \cellcolor{iodnew}68.0 & \cellcolor{iodnew}77.4 & \cellcolor{iodnew}70.2 & \underline{74.5} & \underline{70.2} & \underline{72.3} \\
Decoupled PROB~\cite{inoue2025decoupled} & 75.8 & 74.1 & 67.8 & 54.6 & 44.8 & 76.9 & 82.9 & 83.3 & 46.1 & 70.3 & \cellcolor{iodnew}62.9 & \cellcolor{iodnew}71.3 & \cellcolor{iodnew}72.4 & \cellcolor{iodnew}75.0 & \cellcolor{iodnew}73.9 & \cellcolor{iodnew}36.8 & \cellcolor{iodnew}66.1 & \cellcolor{iodnew}64.7 & \cellcolor{iodnew}74.5 & \cellcolor{iodnew}68.3 & 67.7 & 66.6 & 67.1 \\
\cmidrule(r){1-24}
\rowcolor{oursblue} \textbf{NEST (ours)} & 81.4 & 75.2 & 79.4 & 65.7 & 66.7 & 80.8 & 88.7 & 81.6 & 49.7 & 81.5 & \cellcolor{iodnew}60.0 & \cellcolor{iodnew}74.3 & \cellcolor{iodnew}82.8 & \cellcolor{iodnew}80.1 & \cellcolor{iodnew}81.4 & \cellcolor{iodnew}41.5 & \cellcolor{iodnew}72.2 & \cellcolor{iodnew}66.3 & \cellcolor{iodnew}76.2 & \cellcolor{iodnew}69.0 & \textbf{75.1} & \textbf{70.4} & \textbf{72.7} \\
\midrule
\textbf{15+5} & aero & cycle & bird & boat & bottle & bus & car & cat & chair & cow & table & dog & horse & bike & person & plant & sheep & sofa & train & tv & \textbf{Old} & \textbf{New} & \textbf{mAP} \\
\midrule
ORE~\cite{joseph2021towards} & 75.4 & 81.0 & 67.1 & 51.9 & 55.7 & 77.2 & 85.6 & 81.7 & 46.1 & 76.2 & 55.4 & 76.7 & 86.2 & 78.5 & 82.1 & \cellcolor{iodnew}32.8 & \cellcolor{iodnew}63.6 & \cellcolor{iodnew}54.7 & \cellcolor{iodnew}77.7 & \cellcolor{iodnew}64.6 & 71.8 & 58.7 & 68.5 \\
OW-DETR~\cite{gupta2022owdetr} & 77.1 & 76.5 & 69.2 & 51.3 & 61.3 & 79.8 & 84.2 & 81.0 & 49.7 & 79.6 & 58.1 & 79.0 & 83.1 & 67.8 & 85.4 & \cellcolor{iodnew}33.2 & \cellcolor{iodnew}65.1 & \cellcolor{iodnew}62.0 & \cellcolor{iodnew}73.9 & \cellcolor{iodnew}65.0 & 72.2 & 59.8 & 69.4 \\
PROB~\cite{zohar2023prob} & 77.9 & 77.0 & 77.5 & 56.7 & 63.9 & 75.0 & 85.5 & 82.3 & 50.0 & 78.5 & 63.1 & 75.8 & 80.0 & 78.3 & 77.2 & \cellcolor{iodnew}38.4 & \cellcolor{iodnew}69.8 & \cellcolor{iodnew}57.1 & \cellcolor{iodnew}73.7 & \cellcolor{iodnew}64.9 & 73.2 & 60.8 & 70.1 \\
CAT~\cite{ma2023cat} & 75.3 & 81.0 & 84.4 & 64.5 & 56.6 & 74.4 & 84.1 & 86.6 & 53.0 & 70.1 & 72.4 & 83.4 & 85.5 & 81.6 & 81.0 & \cellcolor{iodnew}32.0 & \cellcolor{iodnew}58.6 & \cellcolor{iodnew}60.7 & \cellcolor{iodnew}81.6 & \cellcolor{iodnew}63.5 & 75.6 & 59.3 & 72.2 \\
OrthogonalDet~\cite{sun2024exploring} & 82.7 & 80.4 & 78.5 & 55.3 & 65.5 & 81.0 & 89.8 & 85.9 & 52.6 & 84.6 & 62.3 & 78.4 & 82.7 & 81.1 & 84.2 & \cellcolor{iodnew}46.5 & \cellcolor{iodnew}71.6 & \cellcolor{iodnew}79.0 & \cellcolor{iodnew}82.5 & \cellcolor{iodnew}79.2 & \underline{76.3} & \textbf{71.8} & \underline{74.7} \\
Decoupled PROB~\cite{inoue2025decoupled} & 79.6 & 77.0 & 74.9 & 57.8 & 62.8 & 77.0 & 84.9 & 84.4 & 46.4 & 73.8 & 64.9 & 76.4 & 80.6 & 77.6 & 76.7 & \cellcolor{iodnew}41.8 & \cellcolor{iodnew}72.6 & \cellcolor{iodnew}61.1 & \cellcolor{iodnew}74.5 & \cellcolor{iodnew}65.4 & 73.0 & 63.1 & 70.5 \\
\cmidrule(r){1-24}
\rowcolor{oursblue} \textbf{NEST (ours)} & 83.5 & 81.6 & 77.0 & 61.5 & 68.4 & 82.6 & 89.8 & 85.1 & 45.0 & 86.5 & 65.0 & 86.6 & 84.5 & 84.9 & 83.1 & \cellcolor{iodnew}40.9 & \cellcolor{iodnew}73.4 & \cellcolor{iodnew}68.9 & \cellcolor{iodnew}79.8 & \cellcolor{iodnew}74.0 & \textbf{77.7} & \underline{67.4} & \textbf{75.1} \\
\midrule
\textbf{19+1} & aero & cycle & bird & boat & bottle & bus & car & cat & chair & cow & table & dog & horse & bike & person & plant & sheep & sofa & train & tv & \textbf{Old} & \textbf{New} & \textbf{mAP} \\
\midrule
ORE~\cite{joseph2021towards} & 67.3 & 76.8 & 60.0 & 48.4 & 58.8 & 81.1 & 86.5 & 75.8 & 41.5 & 79.6 & 54.6 & 72.8 & 85.9 & 81.7 & 82.4 & 44.8 & 75.8 & 68.2 & 75.7 & \cellcolor{iodnew}60.1 & 69.4 & 60.1 & 68.9 \\
OW-DETR~\cite{gupta2022owdetr} & 70.5 & 77.2 & 73.8 & 54.0 & 55.6 & 79.0 & 80.8 & 80.6 & 43.2 & 80.4 & 53.5 & 77.5 & 89.5 & 82.0 & 74.7 & 43.3 & 71.9 & 66.6 & 79.4 & \cellcolor{iodnew}62.0 & 70.2 & 62.0 & 70.2 \\
PROB~\cite{zohar2023prob} & 80.3 & 78.9 & 77.6 & 59.7 & 63.7 & 75.2 & 86.0 & 83.9 & 53.7 & 82.8 & 66.5 & 82.7 & 80.6 & 83.8 & 77.9 & 48.9 & 74.5 & 69.9 & 77.6 & \cellcolor{iodnew}48.5 & 73.9 & 48.5 & 72.6 \\
CAT~\cite{ma2023cat} & 86.0 & 85.8 & 78.8 & 65.3 & 61.3 & 71.4 & 84.8 & 84.8 & 52.9 & 78.4 & 71.6 & 82.7 & 83.8 & 81.2 & 80.7 & 43.7 & 75.9 & 58.5 & 85.2 & \cellcolor{iodnew}61.1 & 74.4 & 61.1 & 73.8 \\
OrthogonalDet~\cite{sun2024exploring} & 83.8 & 84.7 & 77.0 & 62.9 & 60.8 & 80.9 & 88.6 & 85.8 & 51.1 & 81.4 & 67.2 & 86.7 & 86.3 & 83.4 & 83.4 & 44.7 & 74.5 & 73.1 & 81.1 & \cellcolor{iodnew}74.9 & \textbf{75.7} & \textbf{74.9} & \textbf{75.6} \\
Decoupled PROB~\cite{inoue2025decoupled} & 81.6 & 78.4 & 76.5 & 58.4 & 64.9 & 77.0 & 85.3 & 83.2 & 49.6 & 80.9 & 67.2 & 82.0 & 83.5 & 76.8 & 79.9 & 46.3 & 74.7 & 70.4 & 84.8 & \cellcolor{iodnew}62.8 & 73.8 & 62.8 & 73.2 \\
\cmidrule(r){1-24}
\rowcolor{oursblue} \textbf{NEST (ours)} & 83.7 & 81.8 & 75.1 & 64.7 & 59.9 & 78.6 & 89.8 & 80.5 & 50.2 & 86.4 & 65.2 & 79.6 & 85.6 & 82.9 & 82.2 & 45.7 & 73.3 & 65.6 & 83.8 & \cellcolor{iodnew}72.5 & \underline{74.5} & \underline{72.5} & \underline{74.3} \\
\bottomrule
\end{tabular}
\caption{\textbf{Incremental object detection on PASCAL VOC 2007, per class.} AP@50 after the last stage. Higher is better. \emph{Old} is the mean AP over the classes learned in the first stage. \emph{New} is the mean AP over the classes learned in the second stage, shaded in gray. \emph{mAP} is the mean AP over all classes. Bold: best, underline: 2nd best.}
\label{tab:supp-iod}
\end{table*}

\subsection{Unknown Object Categorization}
\label{sec:supp-cat}

In \cref{sec:categorization}, we evaluate all categorizers only on the features of our detector, since it matches 1.8 to 4.0 times as many unknown test objects as RandBox or OrthogonalDet at Tasks~1--3 of R-OWODB (\cref{tab:supp-found}).
Its matched objects also include 91.8\%, 83.7\% and 84.0\% of those matched by RandBox or OrthogonalDet, so most objects that the other two detectors match are evaluated as well.
Our categorizer is detector-agnostic, so here we also evaluate all five categorizers on the features of RandBox and OrthogonalDet.

\begin{table}[htbp]
\centering
\footnotesize
\begin{tabular}{lccc}
\toprule
Detector & Task 1 & Task 2 & Task 3 \\
\midrule
RandBox~\cite{wang2023randbox} & 1798 & 1135 & 662 \\
OrthogonalDet~\cite{sun2024exploring} & 2463 & 1549 & 1191 \\
\cmidrule(r){1-4}
\rowcolor{oursblue} NEST (ours) & 7259 & 3899 & 2134 \\
\bottomrule
\end{tabular}
\caption{\textbf{Matched unknown objects on R-OWODB.} The unknown test objects matched to a prediction labeled unknown, on which the categorizers are evaluated.}
\label{tab:supp-found}
\end{table}

\textbf{Baselines.}
To compare fairly, all five categorizers learn from the same annotated known objects, and none of the baselines is tuned on unknown objects.
The known objects come from the 75650 training images at Task~1 and from the 1588 and 2378 fine-tuning images at Tasks~2 and~3.
Each categorizer trains on a random 90\% of these images, and the other 10\% serve as validation images.
HND~\cite{lee2018hierarchical}, ProHOC~\cite{wallin2025prohoc}, HSC~\cite{linderman2023finegrain} and DARTS~\cite{deng2012hedging} follow their released recipes.
ProHOC keeps the epoch with the highest validation accuracy and decides by the expected tree distance.
HSC trains with the two loss settings of its released code and keeps the one with the higher validation accuracy.
DARTS chooses the regularization of its linear \glspl{svm} by validation accuracy and sets its trade-off between accuracy and specificity to guarantee 90\% accuracy on the validation images.
The same training applies to the features of RandBox and OrthogonalDet.

\textbf{Metrics.}
In the COCO taxonomy, groups exist at only two levels, the root and the retained supercategories, so the \gls{auc} measures the trade-off between them.
Every categorizer scores the root and each retained supercategory for a matched object.
For a baseline, these scores are the values that its own decision rule compares, such as the novel outputs of HND.
We add an offset to the root score and assign the root when this sum is at least the highest supercategory score.
As the offset rises from $-\infty$ to $+\infty$, the share of root-target objects assigned to the root rises from 0 to 1, and the share of supercategory-target objects assigned to their target falls to 0.
The \gls{auc} is the area under the curve of the second share against the first.
At a zero offset, each categorizer makes its own assignments, on which Node-exact is measured.
\Gls{e2e} counts the same correct assignments but divides them by all unknown test objects of the task instead of the matched ones, so an object that the detector misses counts as an error.
Node-exact and \gls{auc} therefore compare categorizers only on the matched objects of one detector, while \gls{e2e} also compares across detectors, and all three are computed for each unknown class and then averaged over the classes.

\begin{table*}[htbp]
\centering
\footnotesize
\setlength{\tabcolsep}{4.44pt}
\setlength{\aboverulesep}{0.25ex}\setlength{\belowrulesep}{0.35ex}
\begin{tabular}{lcccccccc}
\toprule
 & \multicolumn{3}{c}{Task 1} & \multicolumn{3}{c}{Task 2} & \multicolumn{2}{c}{Task 3} \\
\cmidrule(lr){2-4}\cmidrule(lr){5-7}\cmidrule(l){8-9}
Categorizer & NE$\uparrow$ & AUC$\uparrow$ & E2E$\uparrow$ & NE$\uparrow$ & AUC$\uparrow$ & E2E$\uparrow$ & NE$\uparrow$ & E2E$\uparrow$ \\
\midrule
\multicolumn{9}{l}{\textit{Detector: RandBox}~\cite{wang2023randbox}} \\
\quad HND~\cite{lee2018hierarchical} & 58.95\,$\pm$\,0.00 & \textbf{61.93}\,$\pm$\,0.33 & 8.56\,$\pm$\,0.00 & 38.85\,$\pm$\,0.31 & \underline{33.92}\,$\pm$\,0.20 & 7.42\,$\pm$\,0.02 & 39.68\,$\pm$\,0.16 & 9.67\,$\pm$\,0.06 \\
\quad ProHOC~\cite{wallin2025prohoc} & 63.69\,$\pm$\,0.16 & 59.49\,$\pm$\,0.87 & 9.84\,$\pm$\,0.08 & 35.18\,$\pm$\,0.46 & 30.68\,$\pm$\,0.54 & 5.06\,$\pm$\,0.53 & 26.29\,$\pm$\,1.91 & 6.24\,$\pm$\,0.65 \\
\quad HSC~\cite{linderman2023finegrain} & \underline{63.93}\,$\pm$\,0.18 & 60.92\,$\pm$\,0.19 & \underline{10.13}\,$\pm$\,0.04 & \underline{40.17}\,$\pm$\,0.85 & 33.28\,$\pm$\,0.68 & \underline{7.57}\,$\pm$\,0.04 & \underline{41.69}\,$\pm$\,0.11 & \textbf{10.14}\,$\pm$\,0.06 \\
\quad DARTS~\cite{deng2012hedging} & 63.20 & 57.12 & 9.91 & 37.02 & 26.99 & 6.90 & 38.06 & 9.21 \\
\cmidrule(r){1-9}
\rowcolor{oursblue} \quad NEST (ours) & \textbf{65.50}\,$\pm$\,0.39 & \underline{61.51}\,$\pm$\,0.24 & \textbf{10.26}\,$\pm$\,0.03 & \textbf{41.73}\,$\pm$\,0.22 & \textbf{34.90}\,$\pm$\,0.11 & \textbf{7.74}\,$\pm$\,0.01 & \textbf{42.14}\,$\pm$\,0.39 & \underline{9.99}\,$\pm$\,0.01 \\
\midrule
\multicolumn{9}{l}{\textit{Detector: OrthogonalDet}~\cite{sun2024exploring}} \\
\quad HND~\cite{lee2018hierarchical} & 52.48\,$\pm$\,0.05 & \textbf{61.27}\,$\pm$\,0.01 & 9.98\,$\pm$\,0.01 & \textbf{41.42}\,$\pm$\,0.47 & \textbf{35.65}\,$\pm$\,0.35 & 9.66\,$\pm$\,0.04 & 40.14\,$\pm$\,0.76 & 15.99\,$\pm$\,0.15 \\
\quad ProHOC~\cite{wallin2025prohoc} & 61.56\,$\pm$\,0.12 & 57.78\,$\pm$\,0.49 & 12.50\,$\pm$\,0.04 & 32.94\,$\pm$\,0.66 & 29.65\,$\pm$\,0.93 & 6.95\,$\pm$\,0.08 & 24.43\,$\pm$\,1.55 & 10.08\,$\pm$\,0.70 \\
\quad HSC~\cite{linderman2023finegrain} & \underline{64.13}\,$\pm$\,0.41 & 58.86\,$\pm$\,0.21 & \underline{12.98}\,$\pm$\,0.12 & \underline{40.88}\,$\pm$\,0.11 & 33.52\,$\pm$\,0.06 & \underline{9.80}\,$\pm$\,0.03 & \underline{46.01}\,$\pm$\,0.42 & \underline{17.05}\,$\pm$\,0.07 \\
\quad DARTS~\cite{deng2012hedging} & 61.02 & 52.22 & 12.41 & 35.59 & 27.24 & 8.66 & 40.55 & 15.45 \\
\cmidrule(r){1-9}
\rowcolor{oursblue} \quad NEST (ours) & \textbf{65.47}\,$\pm$\,0.27 & \underline{60.50}\,$\pm$\,0.35 & \textbf{13.15}\,$\pm$\,0.07 & 40.13\,$\pm$\,0.23 & \underline{34.35}\,$\pm$\,0.15 & \textbf{9.99}\,$\pm$\,0.02 & \textbf{48.18}\,$\pm$\,0.43 & \textbf{17.46}\,$\pm$\,0.06 \\
\midrule
\multicolumn{9}{l}{\textit{Detector: NEST (ours)}} \\
\quad HND~\cite{lee2018hierarchical} & 56.35\,$\pm$\,0.02 & 58.58\,$\pm$\,0.03 & 27.61\,$\pm$\,0.02 & \underline{36.52}\,$\pm$\,0.17 & \underline{30.71}\,$\pm$\,0.07 & 15.49\,$\pm$\,0.20 & 33.10\,$\pm$\,0.13 & 17.76\,$\pm$\,0.15 \\
\quad ProHOC~\cite{wallin2025prohoc} & 58.91\,$\pm$\,0.65 & 56.54\,$\pm$\,0.08 & 28.95\,$\pm$\,0.48 & 27.28\,$\pm$\,1.06 & 25.18\,$\pm$\,0.86 & 9.84\,$\pm$\,0.89 & 17.82\,$\pm$\,0.85 & 9.92\,$\pm$\,0.51 \\
\quad HSC~\cite{linderman2023finegrain} & \underline{64.59}\,$\pm$\,0.30 & \underline{59.26}\,$\pm$\,0.76 & \underline{32.20}\,$\pm$\,0.13 & 34.58\,$\pm$\,1.19 & 28.11\,$\pm$\,1.25 & \underline{16.26}\,$\pm$\,0.31 & \underline{45.43}\,$\pm$\,1.60 & \underline{24.40}\,$\pm$\,0.71 \\
\quad DARTS~\cite{deng2012hedging} & 62.12 & 54.20 & 30.93 & 29.68 & 21.62 & 13.79 & 38.22 & 21.44 \\
\cmidrule(r){1-9}
\rowcolor{oursblue} \quad NEST (ours) & \textbf{64.98}\,$\pm$\,0.07 & \textbf{59.49}\,$\pm$\,0.65 & \textbf{32.22}\,$\pm$\,0.06 & \textbf{36.89}\,$\pm$\,0.26 & \textbf{31.83}\,$\pm$\,0.37 & \textbf{17.39}\,$\pm$\,0.09 & \textbf{49.51}\,$\pm$\,0.28 & \textbf{26.65}\,$\pm$\,0.20 \\
\bottomrule
\end{tabular}
\caption{\textbf{Unknown categorization on R-OWODB with the features of each detector.} Node-exact (NE), AUC and E2E in \%, as mean $\pm$ standard deviation over the three training runs of each categorizer. DARTS is deterministic. Bold: best, underline: 2nd best.}
\label{tab:supp-seed-sd}
\end{table*}

\textbf{Results.}
In \cref{tab:supp-seed-sd}, our categorizer has the highest Node-exact in eight of the nine cases across the three detectors.
The exception is OrthogonalDet at Task~2, where HND and HSC score higher.
Its \gls{auc}, defined only at Tasks~1 and~2, is the best or second best in all six cases.
Its lead therefore holds on the features of every detector, not only on ours.
With the features of RandBox and OrthogonalDet, however, no categorizer reaches an \gls{e2e} above 13.2\%, 10.0\% and 17.5\% at Tasks~1--3, while our detector and categorizer together reach 1.5 to 2.4 times these values (\cref{tab:random}).
Both parts contribute, since our detector matches many more unknown objects (\cref{tab:supp-found}) and our categorizer has the highest \gls{e2e} in eight of the nine cases.

\Cref{tab:supp-seed-sd} also gives the standard deviation over three training runs with seeds 0, 1 and 2 on the same detector checkpoint, for every categorizer except the deterministic DARTS.
These have a median of 0.3 points for Node-exact and \gls{auc} and 0.07 points for \gls{e2e}.
ProHOC and HSC vary most, by up to 1.9 points, while our categorizer varies by at most 0.65 points.

\begin{figure}[htbp]
  \centering
  \captionsetup[sub]{justification=raggedright}
  \begin{subfigure}{\linewidth}
    \centering
    \includegraphics{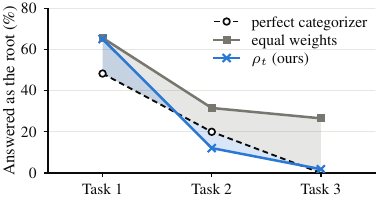}
    \caption{Share of matched unknown objects whose target is the root, which a perfect categorizer would reproduce, and share that our categorizer assigns to the root with equal weights or with $\rho_t$. Closer to the dashed line is better, and shading marks the gap.}
  \end{subfigure}\par\medskip
  \begin{subfigure}{\linewidth}
    \centering
    \includegraphics{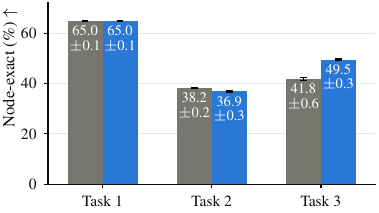}
    \caption{Node-exact, mean $\pm$ standard deviation over three training runs. The largest effect is at Task~3, where $\rho_t$ removes nearly all root answers and raises Node-exact by 7.7 points.}
  \end{subfigure}
  \caption{\textbf{Effect of $\rho_t$ on R-OWODB with our detector.}}
  \label{fig:supp-rho}
\end{figure}

\textbf{Effect of $\rho_t$.}
To test the cut weighting of \cref{sec:family_geometry}, we also train our categorizer with equal weights for the cuts at different levels, which hide a class alone or together with its group (\cref{fig:supp-rho}).
Weighting the cuts by $\rho_t$ calibrates the level of the categorizer's answers to the level of the targets of unknown objects, which $\rho_t$ estimates from the known classes.
With the two levels of the COCO taxonomy, the weighting sets how often the categorizer assigns an object to the root (\cref{fig:supp-rho}a).
With $\rho_t$, this share follows the share of matched objects whose target is the root, with 65.0\%, 12.1\% and 1.9\% at Tasks~1--3 against 48.3\%, 20.0\% and 0.0\%.
With equal weights, it is 65.6\%, 31.6\% and 26.5\%.
The difference is largest at Task~3, where no unknown object has the root as its target.
This is also reflected in Node-exact, which $\rho_t$ raises from 41.8 to 49.5 at Task~3 (\cref{fig:supp-rho}b), while the two weightings differ by at most 1.3 points at Tasks~1 and~2.

\section{More Qualitative Results}
\label{sec:supp-qual}

\begin{figure*}[htbp]
  \centering
  \includegraphics[width=0.88\linewidth]{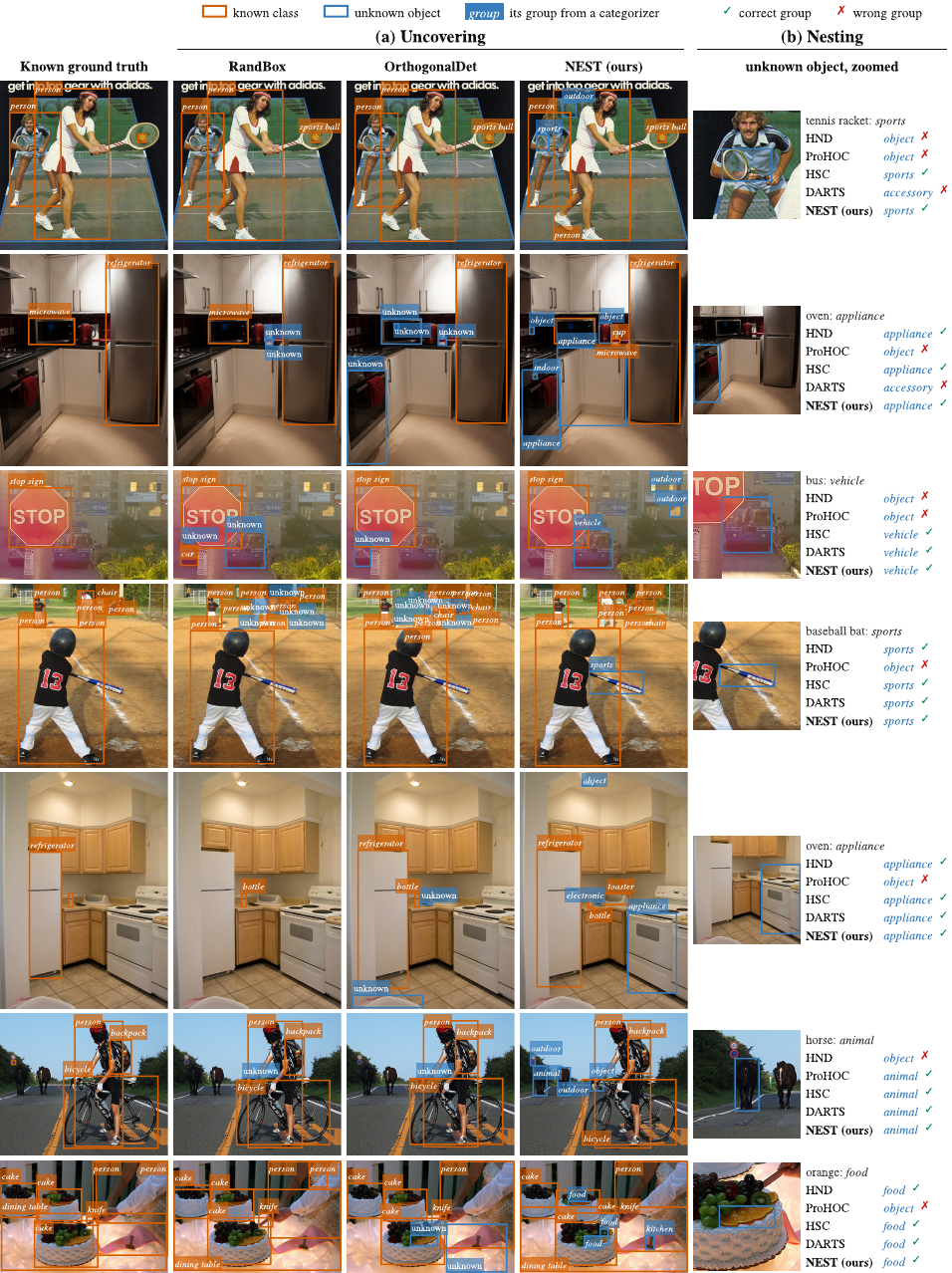}
  \caption{\textbf{More qualitative results on R-OWODB Task~3.} Layout as in \cref{fig:qualitative}. (a)~NEST finds unknown objects that RandBox and OrthogonalDet do not show, such as the baseball bat in the fourth row and the distant horses in the sixth row. (b)~HND and ProHOC often fall back to the root \emph{object}, which is never the target at this task, and DARTS assigns the tennis racket and the oven in the second row to \emph{accessory}. Where they fail, NEST names the target group, such as \emph{sports} for the racket and \emph{vehicle} for the bus.}
  \label{fig:supp-qual}
\end{figure*}

\Cref{fig:supp-qual} shows seven more images of R-OWODB Task~3 in the layout of \cref{fig:qualitative}.
In the first, fourth and sixth rows, our detector finds the tennis racket, the baseball bat and the two distant horses, which RandBox and OrthogonalDet do not show.
RandBox and OrthogonalDet do find some unknown objects, such as the oven in the second row and the buses in the third row.
Our detector also finds objects outside the 80 COCO classes, which carry no annotation, such as the kiwi slice and the cake server in the last row, and our categorizer places them in fitting groups, \emph{food} and \emph{kitchen}.

Part~(b) compares the groups that the five categorizers assign to the same detection from our detector.
At Task~3, no unknown class has the root \emph{object} as its target.
Nevertheless, HND answers \emph{object} for the tennis racket, the bus and the horse, and ProHOC for the racket, the bus, the two ovens, the baseball bat and the orange.
ProHOC answers \emph{object} this often because it chooses the group with the smallest expected tree distance.
With groups at only two levels, this group is a supercategory only when the supercategory and its classes hold more than half of the predicted probability, which no supercategory reaches for these objects.
DARTS makes a different mistake.
It rarely answers \emph{object} but assigns an object to the supercategory of the known class it resembles most.
For the racket, this is the backpack, with 84\% of the probability, so DARTS answers \emph{accessory}, as it does for the oven in the second row.
Our categorizer and HSC, in contrast, assign these objects to their target groups, such as \emph{sports} for the racket and \emph{appliance} for the ovens, in line with their leading performance in \cref{tab:random}.

\clearpage

\clearpage

\end{document}